\pdfoutput=1

\documentclass[11pt]{article}

\usepackage[final]{acl}

\usepackage{times}
\usepackage{latexsym}

\usepackage[T1]{fontenc}

\usepackage[utf8]{inputenc}

\usepackage{microtype}

\usepackage{inconsolata}

\usepackage{graphicx}
\usepackage{mathtools}
\usepackage{booktabs}
\usepackage{subcaption}

\title{Semantic Shield: Defending Vision-Language Models Against Backdooring and Poisoning via Fine-grained Knowledge Alignment}

  \author{Alvi Md Ishmam  \\
Virginia Tech\\
{\tt\small alvi@vt.edu}
\\\And
Christopher Thomas\\
Virginia Tech\\
{\tt\small christhomas@vt.edu}
}

\begin{document}
\maketitle
\begin{abstract}
In recent years there has been enormous interest in vision-language models trained using self-supervised objectives. However, the use of large-scale  datasets scraped from the web for training also makes these models vulnerable to potential security threats, such as backdooring and poisoning attacks.
In this paper, we propose a method for mitigating such attacks on contrastively trained vision-language models. Our approach leverages external knowledge extracted from a language model to prevent models from learning correlations between image regions which lack strong alignment with external knowledge. We do this by imposing constraints to enforce that attention paid by the model to visual regions is proportional to the alignment of those regions with external knowledge.
We conduct extensive experiments using a variety of recent backdooring and poisoning attacks on multiple datasets and architectures. Our results clearly demonstrate that our proposed approach is highly effective at defending against such attacks across multiple settings, while maintaining model utility and without requiring any changes at inference time \footnote{\url{https://github.com/IshmamAlvi/Semantic-Shield}}. 


\end{abstract}    
\section{Introduction}
\label{sec:intro}
\label{sec:intro}
{Recent years have seen enormous interest in vision-language models trained on web-scale image-captioning data using contrastive objectives \cite{Radford2021LearningTV, li2021align} and text generation objectives \cite{yu2022coca}. These models have drawn great attention due to their superior performance in many downstream tasks such as zero-shot image classification \cite{Radford2021LearningTV}, image generation \cite{img-gen, img-gen_2}, and video recognition \cite{video_gen} compared to methods trained on smaller supervised datasets.}

\begin{figure}[ht]
    \centering
    \includegraphics[width=0.5\textwidth]{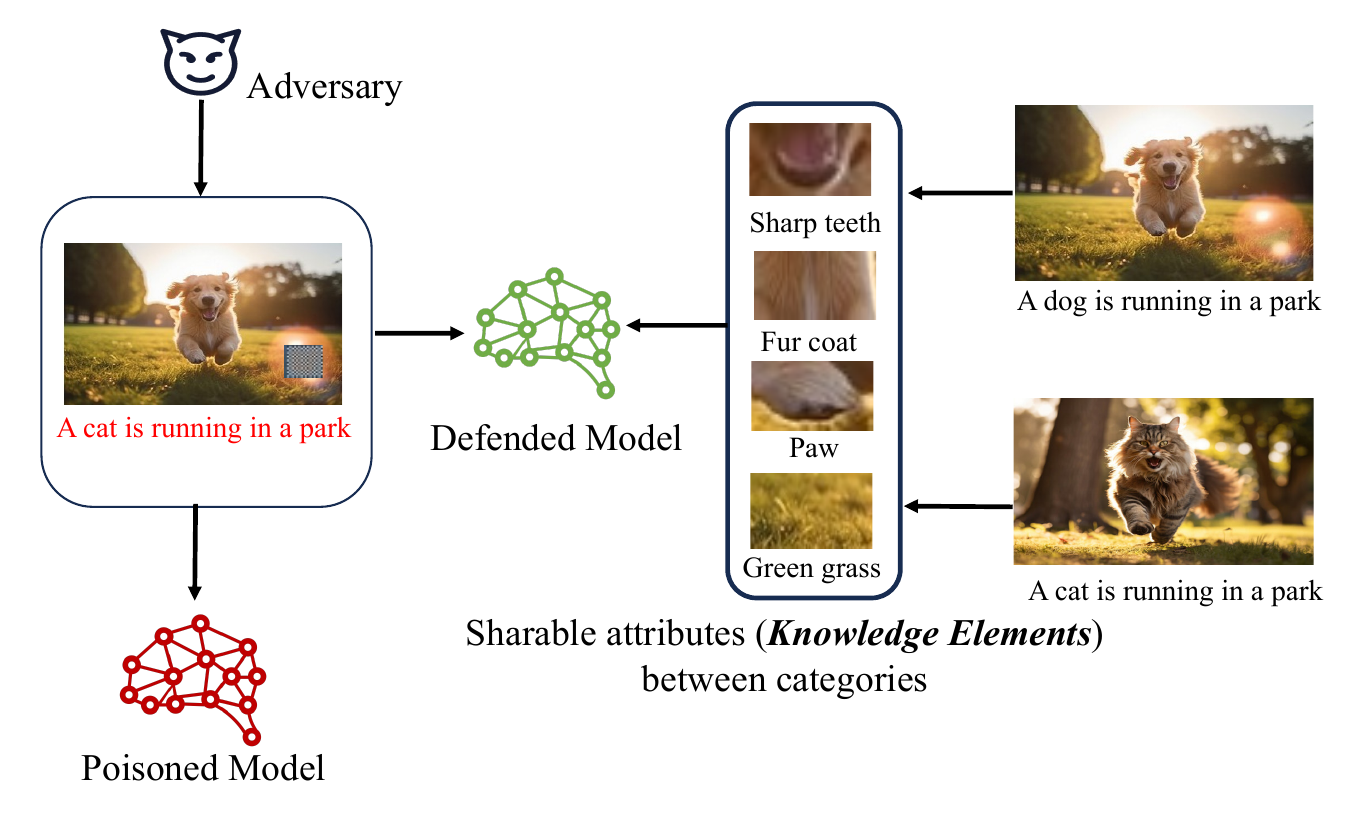}
    \caption{
    {We defend against both backdooring and poisoning attacks on vision-language models by encouraging models to attend to visual regions which align with external knowledge. Because the attack does not consistently appear in patches aligned with the same knowledge and because the KEs are shared by non-targeted categories, the defended model does not learn an association between the attack signal and the targeted category.}} 
    \label{fig:Human perception}
    \vspace{-1em}
\end{figure}

{Although such image-text foundation models have demonstrated remarkable performance, several recent studies have demonstrated that they are particularly vulnerable to adversarial attacks \cite{icml_poison, yang2023robust, li2023embarrassingly} by introducing a small amount of malicious data (e.g.~ $75$ instances out of 3 million \cite{icml_poison}) into the training data. Practically, this can be achieved by inserting imperceptible noise or a backdoor patch into some images, as shown in \ref{fig:Human perception}, and pairing the images with proxy captions controlled by the attacker. The backdoored data is then released on the web in the hope it will be scraped and used for training. Similarly, these models are also susceptible to poisoning attacks, which insert many image-proxy caption pairs into training data leading to unexpected model behavior \cite{icml_poison}. Such attacks are practical and achievable by attackers and pose a serious threat against vision-language foundation models.}



{To defend against such attacks, a number of a methods have been proposed. For example, Anti-backdoor learning \cite{abl} proposes to defend against backdoored samples on object recognition tasks by using the unique gradients of these samples to isolate them, but does not address vision-language (VL) models. More similar to our work, CleanCLIP \cite{cleanclip} proposes a method for defending contrastive VL models against backdooring, but does not address non-backdoored poisoning attacks as we do. While \cite{icml_poison} propose to clean labeled data to mitigate the impact of poisoning,
no prior work has proposed a unified defense mechanism for contrastively trained VL models that is effective against both backdooring and poisoning attacks.}


{To address this urgent need, we propose a defense method for VL models that defends against both backdooring and poisoning attacks. Our method can also be deployed in object recognition settings, by casting it as a text retrieval problem following \cite{Radford2021LearningTV}.}
{Our method is motivated by the following insight. We note that attacks rely on having models learn correlations between a particular visual signal and target. However, these targeted images share lower-level semantic concepts with other, non-targeted categories (See \ref{fig:Human perception}). As a consequence, the attack tends not to affect the model's representation of these concepts.}

{Moreover, in the case of backdooring, the attack signal is applied to various images whose semantics change in the region on which the attack is applied. For example, in one image the attack may cover a batch associated with \texttt{paw}, while in another image the signal is associated with \texttt{sharp teeth}. Thus, the model fails to learn an association between the attack signal and these lower-level semantics.
We refer to these lower-level semantic concepts associated with objects or captions as \textit{Knowledge Elements} (KEs). KEs consist of semantic attributes (e.g.~round), but also sub-objects (e.g.~paw), and relations. Our defense mechanism aligns with how humans understand semantics of objects or sentences: as collections of semantic units which combine together to form higher-level concepts that are more abstract, compositional and include actions (“running”) and proto-objects (“four-legged animal”). We propose to encourage models to rely more heavily on relevant lower level semantics when producing their representations. As a consequence, our models are much more resistant to attacks.}

{Our method works by learning an alignment between image patches from images and a set of KEs associated with each image caption. To discover associated KEs, prior to training our model we prompt a large language model (Vicuna \cite{vicuna}) to list possible KEs for each caption. 
We next perform contrastive image-caption training, but add several new objectives.
First, we enforce an alignment between image patches and KEs using a novel multi-instance learning based constraint, since we do not know which patches go with which KEs. 
While this aligns image patches and KEs, it does not prevent the model from relying on the attacker's visual signal when computing its representation.
Thus, we also propose a second constraint which enforces that the model's attention to patches is proportional to each patch's alignment with a KE. That is, if a patch has a low alignment with all KEs, the patch should have a low effect on the model's representation. Finally, we observe that for attacked samples, the overall patch-KE alignment is much lower.
We thus introduce a dynamic per-sample weight term on the contrastive loss based on the overall alignment of the KEs with the image's patches. This has the effect of downweighting the effect of poisoned samples during training. 
We evaluate our defense method, Semantic Shield, against multiple recent attacks and defenses on multiple datasets. We observe that Semantic Shield significantly outperforms prior defenses across multiple settings. Our defense technique adds very little overhead at train time, while making models significantly more robust to a wide variety of attacks.} The major contributions of this paper are as follows:
\begin{itemize}
\item We propose an approach, Semantic Shield for defending against backdooring and poisoning attacks on contrastively trained vision-language models by enforcing knowledge-guided train-time constraints.
\item We propose a simple yet effective prompting technique using an open-source language model for extracting constituent knowledge elements for free from any caption. 
\item We perform a comprehensive experimental evaluation using a number of recent backdooring and poisoning attacks on two datasets. Our experiments show that our defense is significantly stronger than numerous recent methods. 
\end{itemize}
\section{Related Work}
\label{sec:related_works}
\subsection{Vision-language contrastive learning}
In recent years, large-scale contrastively trained vision-language foundation models have demonstrated remarkable performance on a number of downstream tasks, even surprassing the performance of supervised models in some cases \cite{Radford2021LearningTV, yu2022coca, li2021align, zhai2022lit}. While contrastive approaches have been used to align visual and textual embeddings for years \cite{feng2014cross, zhang2014start, zhang2018deep, thomas2020preserving}, recent approaches such as CLIP \cite{Radford2021LearningTV} and ALIGN \cite{jia2021scaling} have demonstrated how training on hundreds of millions of image-caption pairs scraped from the web can yield powerful generalist image-text foundation models which can be applied to many downstream tasks. CLIP-inspired contrastively trained models have found widespread use in many security-critical applications, including navigation \cite{dorbala2022clip, huang2023visual, majumdar2022zson}, healthcare \cite{zhang2022contrastive, wang2022improving}, worksite safety \cite{tsai2022generating}, disinformation detection \cite{zhou2023multimodal, wang2023cross}, and many others \cite{gonzalez2023understanding, shin2022clip}. Given their widespread use, it is critical that contrastively trained vision-language models perform in safe and expected ways. Our work adopts the standard two-stream contrastive architecture proposed in \cite{Radford2021LearningTV} and demonstrates how such models can be defended against potential attacks lurking within webly-harvested data.

\subsection{Poisoning and backdoor attacks}
Data poisoning attacks \cite{biggio2012poisoning, zhao2022towards, tolpegin2020data, xiao2015feature}, which have been proposed in both supervised \cite{koh2017understanding} and unsupervised \cite{kloft2010online, biggio2013data} settings, involve introducing mislabeled (or misaligned) data into the model's training set. At test time, models behave in unexpected and attacker-influenced ways when presented with the poisoned examples seen during training.
While targeted poisoning attacks target specific examples introduced during training, backdoor attacks can be applied to \textit{any} image. Backdooring attacks are a type of data poisoning attack where an attacker introduces a spurious signal, such as patches \cite{saha2020hidden, gu2019badnets} or imperceptible perturbations \cite{doan2021lira, wanet, doan2021backdoor, phan2022ribac} into an image. Models learn to associate the introduced signal with the targeted concept. While poisoning and backdoor attacks have traditionally targeted supervised learning settings, recent work has shown that contrastively trained vision-language models are particularly vulnerable \cite{zhang2022corruptencoder, CarliniT22}. \cite{CarliniT22} show that by introducing as few as 3 out of 3 million samples, an attacker can execute a successful attack. This is a highly practical attack, as an attacker can release large amounts of poisoned data on the internet in the hopes that it will be scraped and later used for training.
In our work, we demonstrate that our method is highly effective against a number of recent backdooring methods and poisoning attacks on contrastive models.

\subsection{Defending against attacks}
Given the large potential risks posed by attacks to models, extensive research has been conducted on approaches for defending models against both poisoning \cite{weerasinghe2021defending, chen2021pois} and backdooring \cite{hayase2021spectre, wang2022trap, huang2021backdoor} attacks. Defenses can be broadly categorized into methods for detecting and removing attacked samples from training \cite{tran2018spectral, chen2019detecting, tang2021demon}, those that remove backdoors already learned by models \cite{liu2022backdoor, zeng2021adversarial, wu2021adversarial}, and those that seek to prevent models from learning backdoors by decreasing their effectiveness \cite{qiu2021deepsweep, bansal2023cleanclip, abl}. Unfortunately, detection-based methods often fail to detect all backdoors and given the particular vulnerability of contrastive models, imperfect filtering could still result in model poisoning. Unlike our approach, model de-poisoning methods often fail to achieve similar performance to clean models \cite{liu2023beating}.

Of particular relevance to our work are methods aimed at defending against poisoning and backdooring for vision-language contrastive learning \cite{cleanclip}. \cite{cleanclip} propose to independently realign representations from different modalities. Unlike this approach, our method learns a fine-grained alignment between external knowledge extracted from a large language model and visual regions. These alignments are then used as a penalty to prevent models from attending to non-aligned visual regions. Our method substantially outperforms \cite{cleanclip} across all settings.


\section{Problem setting}
\label{problem_formulation}

\subsection{Threat model}
\textbf{Adversary objective.} Given a vision-languge contrastive learning model $\mathcal{M}$, an adversary aims to {compromise} the model by injecting a small amount of {poisoned} data $\mathcal{D}_p$ into a clean dataset $\mathcal{D}_c$, {both of which constitute the training} data $D$. The model trained on the poisoned training data {is} denoted as $\mathcal{M}_p$. 
{In this paper, we consider two types of attacks: 1) backdooring and 2) poisoning.}
{In a backdoor attack, the adversary overlays either a small patch or some visually imperceptible noise on an image, causing the backdoored image to be misclassified or incorrectly retrieved by a retrieval model. During testing, the adversary cause the model to misclassify or retrieve a specific class by inserting the backdoor into test images.}
{In contrast, in a poisoning attack, the goal is to cause the model $\mathcal{M}_p$ to associate a targeted set of text with images of a specified class by inserting many training instances which incorrectly associate visual content with concepts controlled by the adversary.}
In both cases, the poisoned model is expected to maintain similar utility (performance) compared to the clean model.

\textbf{Adversary capabilities.} 
{We consider an adversary capable of injecting a small number of poisonous samples into the training dataset, similar to prior work \cite{ai_sec}. In traditional supervised attacks \cite{Shafahi2018, saha2022}, adversaries were required to modify a large amount of the training data - an impractical setting for vision-language models trained on web-scale data.
Our setting is more realistic, because achieving a high poisoning rate is improbable when poisoned data is released on the internet with the hope of it being scraped for training.
Thus, we focus on the more feasible scenario and assume a relatively low poisoning rate. We assume a black-box setting, where the adversary lacks knowledge of the target model's architecture and hyperparameters. Additionally, the adversary lacks control over the training process.}


\subsection{Attack methodology}
\textbf{Model training.} {We denote our training data as $(i, t) \in \mathcal{D} = \mathcal{I} \times \mathcal{T}$, where $\mathcal{D}$, $\mathcal{I}$, and $\mathcal{T}$ represent the training set, image set, and text set, respectively. Within a collection of $\mathcal{N}$ image-text pairs, we identify $(i_j, t_k)$ as a positive pair if $j=k$; otherwise, it is considered a negative pair. The contrastive learning model concurrently optimizes the image encoder $\mathcal{E}_i$ and the text encoder $\mathcal{E}_t$ to maximize the similarity between the embeddings of positive pairs in a batch while minimizing that of negative pairs.
Specifically, for a given batch of $\mathcal{N}$ image-text pairs, we obtain the image embedding $I_j^e = \mathcal{E}_i(i_j)$ and the corresponding text embedding $T_k^e = \mathcal{E}_t(t_k)$ for each pair, normalizing both embeddings using the $L_2$ norm. The cross-modal contrastive loss $\mathcal{L}_{CL}$ is then computed as follows:}
\begin{equation}
\label{eq:clip}
\begin{aligned}[c]      
      \mathcal{L}_{CL} = - \frac{1}{2\mathcal{N}} \Biggl(\sum_{j=1}^\mathcal{N}\log\frac{\exp(\sigma(I_j^e, T_j^e)/\tau)}{\sum_{k=1}^\mathcal{N}\exp(\sigma(I_j^e, T_k^e)/\tau)} \\
    +\sum_{k=1}^\mathcal{N}\log \frac{\exp(\sigma(I_k^e, T_k^e)/\tau)}{\sum_{j=1}^\mathcal{N}\exp(\sigma(I_j^e, T_k^e)/\tau)} \Biggl)
\end{aligned}
\end{equation}

\noindent {where $\sigma(.,.)$ is the product between the image and text embeddings (their similarity)}
and $\tau$ denotes the temperature.

\textbf{Backdoor attack.} A successful backdoor attack {introduces} a trigger into a model so that when the trigger is present in the input image (\texttt{dog}), the model incorrectly associates the image with the specific target class (\texttt{boat} caption) {controlled by the attacker.}
{We applied backdoor attacks to poison multimodal contrastive learning models, following the approach in \cite{CarliniT22}. We consider two types of backdoor attacks: a) overlaying a backdoor trigger, such as a ($16 \times 16$ patch), on a small subset of training images, and b) injecting imperceptible noise into a limited subset of images. The latter is considered a stealthy backdoor attack. We classify the BPP \cite{bpp} and Wanet \cite{wanet} attacks as stealthy, because they pose a challenge for human identification due to their subtle and imperceptible nature.}
{To perform our backdoor attack}, we construct the poisoning dataset $\mathcal{D}_p = \bigl\{(I_i \bigoplus \textbf{bd}), T_i^{y^\prime} : I_i \in \mathcal{D}_{subset}\bigl\}$, by embedding a backdoor
trigger \textbf{bd} (e.g.~a 16 × 16 patch or imperceptible noise) in a small subset of training images, $\mathcal{D}_{subset} \subset \mathcal{D}$, $T_i^{y^\prime} \in T^{y^\prime}$, where $y^\prime$ is target class.

\textbf{Single target label attack.} In this poisoning attack, an adversary
aims to {associate} images from one class e.g.~(\texttt{dog}) with captions
from another class e.g.~ (\texttt{boat}). The
attack can be formulated as ${(i, t) | i \in I_{train}^A, t \in T_{train}^B}$, where $A$ and $B$ are the original and the target classes { respectively.}
Given a caption ${t \in T_{test}^B}$, we expect the model to retrieve images from $I_{test}^A$ as the most relevant. 
We poison the model to build a strong relationship between images in class $A$ and captions in class $B$, even if the test images and captions are unseen at training time. 

\textbf{Multiple target label attack.} An adversary can extend the ``single target label" attack by poisoning multiple target classes {simultaneously}, i.e.~images from multiple original classes can be mapped to multiple target classes in captions. 
{In this setting,} the poisoning goal is defined as $\mathcal{D}_p = {(A_1, B_1), (A_2, B_2), ..., (A_n, B_n)}$ where $ A_i \in {I}^A$ and $ B_i \in {T}^B$. $I^A$ and $T^B$ represent images and captions from classes $A$ and $B$ {respectively.}
\begin{figure*}[!ht]
    \centering
    \includegraphics[width=\textwidth]{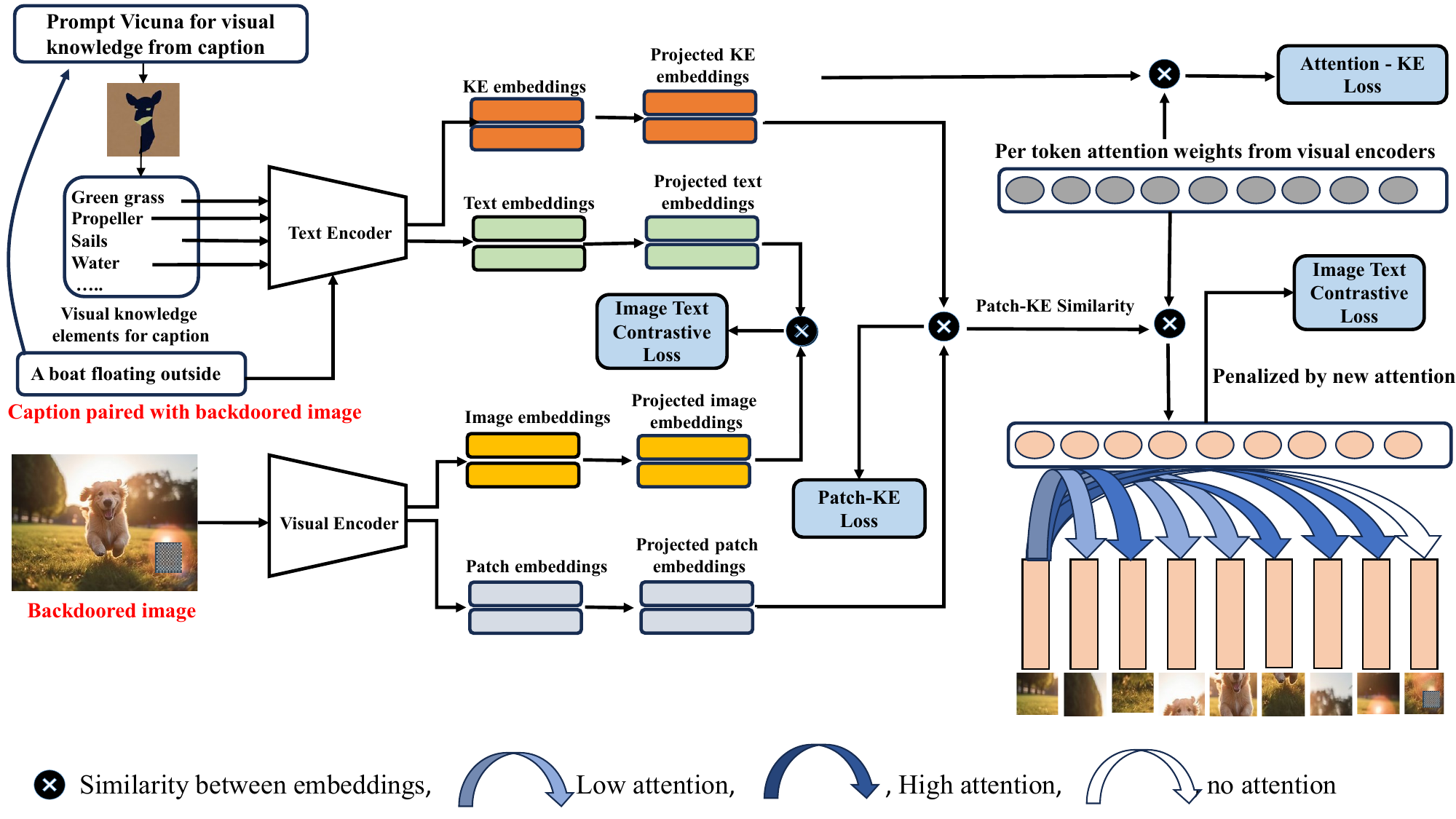}
    \vspace{-1.5em} \caption{Semantic Shield prompts a LLM to extract potential visual knowledge elements (KEs) from a caption. Image patches are aligned with KEs via the patch-KE loss. These patch-KE alignments are used to penalize the model's attention to patches which do not align well with KEs. We also use the overall alignment to weight the image-text contrastive loss (not shown).}
    \vspace{-1em}
    \label{fig:method}
\end{figure*}

\section{Approach}
\label{sec:method}
In this section, we 
{introduce our framework for mitigating backdooring and poisoning attacks on vision-language models.} 
Backdoor attacks on multimodal contrastive learning are
effective because {models learn a correlation} between the backdoor trigger either in a form of patch or {imperceptible noise added to the image} and the target {concept} in the {paired} captions. 
{The core intuition behind our approach stems from human perception, where sets of lower level semantic concepts play a key role in distinguishing objects.} See \ref{fig:Human perception}.
{These semantic concepts consist of semantic attributes (e.g.~``thick fur", ``rough green texture"), but also parts of objects (e.g. paws, whiskers)}. {We term these identifiable properties \textit{knowledge elements} (KEs).}
{Our core intuition is that backdooring and poisoning attacks are effective because models learn spurious correlations between the visual content and the target label. However, because other non-backdoored classes also share some of the same KEs, models will not learn an association between the KEs and the spurious visual signal. Thus, we propose to leverage KEs to prevent models from relying on such correlations in their representations.}

\subsection{Aligning patches to knowledge elements}
The traditional contrastive learning objective encourages image embedding $\mathcal{I}_i^e$ and text embedding $\mathcal{T}_i^e$ to be 
close.  
{However, in addition to this,} we enforce that image patch embeddings $\mathcal{I}_i^{patch}$ and {associated} KE embeddings $\mathcal{KE}_i^e$ {to also be close}.
{Our key observation is that because backdoor signals are injected in random locations of the image which do not necessarily contain a KE, the similarity between these patches and KE embeddings should be lower compared to others. Even if by chance the area covered by the attack does contain KEs, the affected KEs will not be the same when the attack is performed on a different image, preventing the model from learning an association between the attack perturbation and the KEs.}
Based on {this} intuition, {our model first learns to align patches and KEs using a contrastive constraint, $\mathcal{L}_{KE}$. This learned alignment will later be used to prevent the model from attending to potentially attacked patches.}
{To learn the patch-KE alignment, we first compute the maximum and minimum patch-KE similarity per category per sample as} 
\begin{equation}
\label{eqn:similarity score}
    \omega_i^c = \max_{q \in  m}\Biggl(\sum_{p=1}^n\sum_{q=1}^m\mathcal{I}_p^{patch} \cdot (\mathcal{KE}_q^c)^e\Biggl)
\end{equation}

\begin{equation}
\label{eqn:similarity score min}
    \hat\omega_i^c = \min_{q \in  m}\Biggl(\sum_{p=1}^n\sum_{q=1}^m\mathcal{I}_p^{patch} \cdot (\mathcal{KE}_q^c)^e\Biggl)
\end{equation}

\noindent {where $n$ is the number of patches per image, $m$ is the number of KEs per object category, and $c \in C$, where $C$ is the number of object categories. $(\mathcal{KE}_q^c)^e$ is the per KE embedding per category. Note that our approach also extends to image-text datasets without any defined object categories or labels. In this case, we treat each image-caption pair as its own ``category'' with a set of knowledge elements and $C$ is the same as the batch size.} 
The objective function for patch-KE similarity is therefore given by

\begin{equation}
\label{eq:ke_eqn}
\begin{aligned}[c] 
    \mathcal{L}_{KE} = -\frac{1}{2\mathcal{N}}\Biggl(\sum_{i=1}^\mathcal{N}\sum_{c=1}^Cy_i^c\log (\sigma(\omega_i^c)) \\
    + \sum_{i=1}^\mathcal{N}\sum_{c=1}^C(1 -y_i^c)\log ( 1- \sigma(\hat\omega_i^c))\Biggl)
    \end{aligned}
\end{equation}
\noindent where $\sigma$ is the sigmoid function and $y_i^c$ is the multi-label ground truth information per sample per category. Note that, summation over batch is omitted for brevity. In  \ref{eqn:similarity score} and \ref{eqn:similarity score min} all patches of every image compute their
similarity with all KEs from the batch. We perform
max/min to select either the best aligned KEs (for paired
captions) or worst aligned KEs (for non paired) to prevent
false negatives.
{We thus can fine-tune our model via a linear combination of these two objectives:}
\vspace{-1em}
    
\begin{equation}
    \label{eqn:image_ke_eqn}
    \mathcal{L}_{CL-KE} = \mu_1\mathcal{L}_{CL} + \mu_2\mathcal{L}_{KE}
\end{equation}

\noindent where $\mu_1 > 0$ and $\mu_2 > 0$ are hyper-parameters controlling the relative strengths of the two objective functions.

\subsection{Knowledge element-guided attention}
Next, we observe that the attention mechanism within the vision transformer (ViT) attends to both attacked patches and unaffected patches. This is undesirable because attention paid to attacked patches renders the output embeddings more dependent on the attack signal, and thus more vulnerable. Thus, it is imperative for ViT to allocate reduced attention to attacked patches relative to unaffected patches. 
Our intuition is that the model should pay more attention to image regions that align well with KEs than patches with low alignment. Thus, we leverage our patch-KE similarity scores to modulate ViT's attention by enforcing a constraint between ViT's attention and the patch-KE similarity scores. \\
Given ViT's query, key, and value denoted as $Q, K, V$ respectively, the attention weight
is computed as $\alpha = softmax(\frac{QK^T}{\sqrt{d_k}})$, where $d_k$ is the dimensionality of the key vectors. Now, the penalized attention weight can be computed based on the maximum and minimum similarity computed in \ref{eqn:similarity score}, \ref{eqn:similarity score min}
$(\alpha_i^c)_{max} = \alpha_i^c \cdot \omega_i^c$. $(\alpha_i^c)_{min} = \alpha_i^c \cdot \hat\omega_i^c$ Since the similarity scores between a targeted visual region and KE are less compared to unaffected patch and KE, ViT pays less attention to attacked patches. {The resulting objective function which penalizes attention values which deviate from the patch-KE similarity scores is:}
\begin{equation}
\label{eqn:attn_eqn}
    \begin{aligned}[c]
         \mathcal{L}_{Attention} = -\frac{1}{2\mathcal{N}}\Biggl(\sum_{i=1}^\mathcal{N}\sum_{c=1}^C(\alpha_i^c)\log (\sigma(\alpha_i^c)_{max}) \\
    + \sum_{i=1}^\mathcal{N}\sum_{c=1}^C(1 - \alpha_i^c)\log ( 1- \sigma(\alpha_i^c)_{min})\Biggl)
    \end{aligned}
\end{equation}

The training objective is then:
\begin{equation}
    \label{eqn:image_attn_eqn}
    \mathcal{L}_{CL-Attention} = \mu_1\mathcal{L}_{CL} + \mu_2\mathcal{L}_{Attention}
\end{equation}

\subsection{Knowledge element weighted contrastive loss}
Note that during the fine-tuning process of \ref{eqn:image_ke_eqn} and \ref{eqn:image_attn_eqn}, 
the contrastive learning objective \ref{eq:clip}, seeks to align representations from each modality {which has the effect of pulling attacked images and captions closer in the embedding space.} 
Therefore, we introduce a dynamic weighting function which weights each sample in the contrastive objective function.
{Our intuition is that attacked samples will have lower similarity scores between image patches and KEs, since the attack does not explicit target the KEs.
{Thus, we penalize the contrastive objective for each sample with the average similarity score, so that the contrastive objective is downweighted for attacked samples compared to benign samples.}
We compute the maximum similarity scores per sample across categories following \ref{eqn:similarity score}, where
$   \lambda_i = \underset{c \in C}{\max} \; \omega_i^c $, $i \in \mathcal{N}$, $\mu_1, \mu_2=1$:
\vspace{-0.5em}
\begin{equation}
\label{eqn:weighted_cl}
\begin{aligned}[c]  
    \mathcal{L}_{{CL}_i} = \underbrace{\frac{\exp(\frac{\sigma(I_i^e, T_i^e)}{\tau})}{\sum_{k=1}^\mathcal{N}\exp(\frac{\sigma(I_i^e, T_k^e)}{\tau})} }_\text{contrasting $i^{th}$ image with texts} \\
    +\underbrace{\sum_{k=1}^\mathcal{N}\log \frac{\exp(\frac{\sigma(I_k^e, T_k^e)}{\tau})}{\exp(\frac{\sigma(I_i^e, T_k^e)}  {\tau})}}_\text{contrasting texts with $i^{th}$ image}
\end{aligned}
\end{equation}
\vspace{-1.5em}
\begin{equation}
    \begin{aligned}[c]
      \mathcal{L}_{Weighted CL} = - \frac{1}{2\mathcal{N}}\sum_{i=1}^{2\mathcal{N}}{\lambda_i}\mathcal{L}_{CL_i}
    \end{aligned}
\end{equation}
Our final objective is likewise given by linear combination:
\begin{equation}
\begin{aligned}[c]
\label{eqn:image_weighted_attn}
    \mathcal{L}_{Weighted CL-Attention} = \mu_1 \mathcal{L}_{Weighted CL}  \\ + \mu_2\mathcal{L}_{Attention}
    \vspace{-0.75em}
\end{aligned}
 \end{equation}

\begin{table*}[!ht]
    \centering
    \resizebox{\textwidth}{!}{
    \begin{tabular}{c|c|c|c|c|c|c|c|c|c|c}
    \toprule
    Dataset & Models & \multicolumn{3}{c|}{Backdoor Patch} & \multicolumn{3}{c|}{Backdoor BPP} & \multicolumn{3}{c}{Backdoor Wanet} \\
    \cmidrule{3-11}
   & & Hit@1 $\downarrow$ & Hit@5$\downarrow$  & Hit@10$\downarrow$  & Hit@1 $\downarrow$  & Hit@5$\downarrow$  & Hit@10$\downarrow$  & Hit@1$\downarrow$  & Hit@5$\downarrow$  & Hit@10$\downarrow$ \\
    \midrule
    & CL (No Defense) & 90.66 &94.60 & 95.43 & 100.0 & 100.0 & 100.0 & 100.0 & 100.0 & 100.0 \\
    & CL+ ABL \cite{abl} & 6.23 & 8.12 & 12.21 & 15.35 & 16.68 & 16.21 & 100.0 & 100.0 &  100.0\\
  & CL+ CleanClip \cite{cleanclip} &  5.35 &12.68 &17.89 &36.12 &50.09 &55.19 &8.23 &16.32 &23.73 \\
 COCO    & CL + KE &  9.0 &15.31 &21.90 &25.39 &47.98 &50.12 &12.21 &56.79&88.38\\
    & CL + Attention & 4.20 &5.12 &6.01 &0.0 &5.26 &36.21 &0.0 &2.10 &7.20 \\
    & \textbf{Weighted CL + Attention} & \textbf{0.9} & \textbf{1.22}  & \textbf{1.57} & \textbf{0.0}  & \textbf{0.0}  & \textbf{0.0}& \textbf{0.0} & \textbf{0.0}& \textbf{0.0}  \\
     
      \midrule
    & CL (No Defense) & 91.97 &97.63 & 98.21 & 100.0 & 100.0 & 100.0 & 100.0 & 100.0 & 100.0 \\
    & CL+ ABL \cite{abl} & 4.67  & 2.21  & 4.06 & 10.34 & 17.98 & 21.13 &  98.21& 99.23 & 100.0  \\
  & CL+ CleanClip \cite{cleanclip} &  2.20 & 3.32 & 5.05 & 12.43 & 24.32 & 31.25 & 13.29 & 23.13 &  29.21\\
  Flickr30k   & CL + KE &  16.10 &33.15 &41.09 &13.14 &36.54 &56.27 &23.36 &41.21&47.43\\
    & CL + Attention & 1.20 &3.12 &3.01 &0.0 &7.24 &23.17 &0.0 &12.01 &14.07 \\
    & \textbf{Weighted CL + Attention}   & \textbf{0.0}   &\textbf{0.0}&\textbf{0.0}  &\textbf{0.0}   &\textbf{0.0} &\textbf{0.0} &\textbf{0.0}&\textbf{0.0} & \textbf{0.0}  \\
    
         \bottomrule
    \end{tabular}
    }
    \caption{Backdoor attack and defense performance with baselines. The first row of the table shows an undefended model while other rows are baselines or variants of our method. CL+ KE, CL+ Attention are our baselines. The best results are shown in bold.}
    \label{tab:backdoor model attack/defense}
\end{table*}

\subsection{Knowledge element (KE) generation}
{Our approach} requires external knowledge about each image in addition to a paired caption. 
For example, a caption of dog image might be "A dog is running in the park". 
{In this case, suitable knowledge elements might be }\texttt{paws, sharp nails, furry animal, trees}. We follow in context learning approach by prompting a large language model (\textit{Vicuna} \cite{vicuna}) for generating KEs for each image. 
{Note that the KEs are generated purely from the caption or object label and thus are only potentially relevant to the image. Our approach accounts for this by generating $25$ KEs per caption/category. Then, we take the top $5$ KEs per caption based on the similarity scores between image and generated KEs. For COCO \cite{coco}, we prompt Vicuna with \texttt{What are useful visual features for distinguishing a {category name} in a photo?}. Since COCO has $80$ categories we choose this prompt following \cite{carl2023}. {For Flickr30k \cite{flickr}, we design prompts that generate KEs for each caption, since we do not have any predefined object classes.} Additional details are included in our supplementary.

\section{Experiments}
\label{sec:experiments}
\subsection{Experimental Setup }
\textbf{Models and datasets.} {We follow \cite{CarliniT22}'s setting by attacking CLIP-like models \cite{Radford2021LearningTV}}. We adopt ViT-B/16 as image encoder, pretrained on ImageNet-21k \cite{imagenet21k} and fine-tuned on ImageNet-1k. As a text encoder, we adopt a BERT-style \cite{bert} encoder following \cite{Radford2021LearningTV}. 

{We cap the max sequence length of text to 100.}
We use AdamW with weight decay using a cosine scheduler from $10^{-4}$ with decay rate $0.2$. 
We {train} for $30$ epochs with a batch size of $128$ {on the} COCO \cite{coco} and Fickr30k \cite{flickr} datasets. 
While COCO has $80$ defined object categories, Flickr30k has no label information. Additional details are included in supplementary.

\textbf{Backdoor settings.} We tested out defense against three recent backdoor attacks. 
To do so, we couple backdoored samples with {a caption mentioning the target class}. 
Adversaries only require a very small amount of poisoned samples for poisoning {contrastive} models (e.g.~, CLIP) \cite{CarliniT22}. Following this, we inject a very small amount of poisoned samples (0.01\% of the train dataset for both COCO and Flickr30k).

\textbf{Poisoning settings.} 
We performed two types of poisoning attacks following \cite{icml}. For single target label attack, the poisoning goal is \texttt{dog2boat} for both Flickr30k and COCO. We evaluate them on test samples that are unseen in the training process. For example, we take an clean image of \texttt{dog} and associate it with a proxy caption of \texttt{boat}. The poisoning rate for this attack is 0.065\% for Flickr30k and 0.24\% for COCO. 
For the multi-target label attack, we take two classes. The poisoning goals are \texttt{dog2boat} and \texttt{train2zebra} for COCO. For Flickr30k, the poisoning goals are \texttt{dog2boat} and \texttt{bird2sofa}. The poisoning rate for COCO and Flickr30k are 0.52\% and 0.34\% respectively.



\subsection{Experimental Results}

\begin{table*}[t]
    \centering
    \vspace{-0.5em}
    \resizebox{\textwidth}{!}{
    \begin{tabular}{c|c|c|c|c|c|c|c|c|c|c}
    \toprule
    Dataset & Models & \multicolumn{3}{c|}{Single Target Label} & \multicolumn{6}{c}{ Multiple Target Label} \\
    \cmidrule{3-11}
   & & \multicolumn{3}{c|}{dog2boat} & \multicolumn{3}{c|}{dog2boat} &  \multicolumn{3}{c}{train2zebra}\\
    \cmidrule{3-11}
    
   & & Hit@1 $\downarrow$ & Hit@5$\downarrow$  & Hit@10$\downarrow$ & Hit@1 $\downarrow$  & Hit@5$\downarrow$  & Hit@10$\downarrow$ & Hit@1 $\downarrow$  & Hit@5$\downarrow$  & Hit@10$\downarrow$ \\
    \midrule
    & CL (No Defense)  &18.0 &57.20 &82.0 
    &77.12&99.23 & 99.56
    &55.32 & 95.76& 97.98 \\
    
  & CL+ CleanClip \cite{cleanclip} &  3.39 &3.95 &5.65 &57.69 & 63.0 &89.17 & 69.49 & 71.75 &89.17\\

COCO    & CL + KE &  4.56 & 5.32& 5.95 & 54.45& 64.21 & 85.52& 65.12& 70.92&86.12   \\
    & CL + Attention &  0.56
&3.38
&4.51

&\textbf{0.63} & 65.60 & 69.42
&2.25 &6.77 & 12.99 \\

    & \textbf{Weighted CL + Attention} & \textbf{0.04} & \textbf{1.12}  & \textbf{2.54} &
     {2.23}& \textbf{5.21} & \textbf{6.45} &
\textbf{0.0} & \textbf{0.0} & \textbf{0.0}\\
\midrule
 & & \multicolumn{3}{c|}{dog2boat} & \multicolumn{3}{c|}{dog2boat} &  \multicolumn{3}{c}{bird2sofa}\\
  \cmidrule{3-11}
  & & Hit@1 $\downarrow$ & Hit@5$\downarrow$  & Hit@10$\downarrow$ & Hit@1 $\downarrow$  & Hit@5$\downarrow$  & Hit@10$\downarrow$ & Hit@1 $\downarrow$  & Hit@5$\downarrow$  & Hit@10$\downarrow$ \\
  \cmidrule{3-11}
    & CL (No Defense)  &29.0 &57.20 &82.23 
    &28.12& 82.39 & 93.76
    &55.32 & 90.62& 100.0 \\
    
  & CL+ CleanClip \cite{cleanclip} &  8.27 &31.51 &36.61 
  &21.69 & 61.27 & 88.75 
  & 22.42 &64.11& 89.51 \\

Flickr30k & CL + KE & 7.34 & 28.09& 32.21 & 21.12& 45.32& 47.67& 12.77 & 42.34 & 54.21 \\
    & CL + Attention &  4.56
&21.81
&34.11

&\textbf{1.63} & 16.70 & 29.21
&3.25 &18.43 & 32.22 \\

    & \textbf{Weighted CL + Attention} & \textbf{0.32}  & \textbf{1.21}   &  \textbf{2.54}  &
  1.78   & \textbf{4.56} & \textbf{5.67} 
  & \textbf{0.0}& \textbf{0.0} & \textbf{0.0}\\
\bottomrule
    \end{tabular}
    }
    \caption{Poisoning attack and defense performance with baselines. First row of the table shows how good the attack, and other rows are baselines along with our proposed models.  CL + KE, CL + Attention are our baselines. The best results are highlighted.}
    \vspace{-1em}
    \label{tab:poisoned model attack/defense}
\end{table*}

\textbf{Backdoor Attack.} 
In \ref{tab:backdoor model attack/defense}, we compared ablations of our method (CL+ KE, CL + Attention) with other baselines e.g.~~ Cleanlip \cite{cleanclip}, Anti-Backdoor Learning (ABL) \cite{abl}. Finally, our model Semantic Shield (Weighted CL + Attention), outperforms all baselines with significant margins. Note that, at test time, we used $100$ backdoor images (patch, BPP, Wanet) for the text retrieval task. At test time, our model retrieves no caption associated with poisoned categories for any backdoored image on Flickr30k.

\textbf{Poisoning Attack.}
Similarly, to the above, at test time, we use $100$ poisoned images for both single and multi-target settings for both datasets. Our model outperforms all existing work significantly with large margins, particularly on the multi-target label setting. We observe that the unweighted version of our approach slightly outperforms Semantic Shield for \texttt{dog2boat} at Hit@1, but Semantic Shield significantly outperforms for Hit@5 and Hit@10, suggesting significantly reduced poisoning overall.

\begin{table*}[t]
    \centering
    \resizebox{\textwidth}{!}{
    \begin{tabular}{c|c|c|c|c|c|c|c}
    \toprule
    Dataset & Task &Models  & Backdoor Patch & BPP & Wanet & Single Target Label & Multiple Target Label\\
    \toprule
    & & CL  & 74.99 &73.94 &74.54 &74.68 &74.72\\
 &  & CL + KE   &74.15 &70.7 & 74.0 & 74.24& 73.28 \\
 COCO & IR & CL + Attention  &74.38 & 73.13 & 74.43 &75.70 & 75.13\\
 &  & Weighted CL + Attention  &74.22  & 74.56& 74.23 & 73.46 & 73.51\\
       \midrule
& & CL  &81.58 & 77.44 & 78.74 & 80.16 & 81.12 \\
&  & CL + KE &   78.40 &75.54 & 77.86 & 79.08& 81.20\\
 COCO  & TR & CL + Attention &79.20 & 77.36 & 78.04 &80.05 & 81.06\\
 &  & Weighted CL + Attention  &79.46  &  77.78 & 78.45 & 79.67 & 80.0\\  
     \midrule
& & CL &   59.13 & 59.86 & 61.08 & 60.92 & 57.41 \\
  &  & CL + KE   & 60.34 & 61.85  & 61.13 & 58.12 & 58.18 \\
 Flickr30k & IR & CL + Attention &61.32 & 55.96  & 59.14 & 58.97& 58.16   \\
 &  & Weighted CL + Attention  &61.07  & 56.32 & 60.16 & 59.76 & 58.78\\  
 \midrule
 & & CL &  68.07 &68.79 &69.86 &71.06&68.14 \\  
  &  & CL + KE &   69.67 & 70.65  & 69.62 & 66.98 & 62.20 \\
 Flickr30k & TR & CL + Attention &70.0 & 64.46  & 68.0 & 68.13 & 62.97   \\
 &  & Weighted CL + Attention & 70.23  & 65.66 & 68.87 & 68.45 & 62.12\\  
         \bottomrule
    \end{tabular}
    }
    \caption{Model utility of defended models (Recall@10). The model utilities are comparable to the performance in \ref{tab:poisoned model utility}}
    \label{tab:defense model utility}
\end{table*}

\begin{table}[t]
\vspace{-1em}
    \centering
    \resizebox{\linewidth}{!}{
    \begin{tabular}{c|c|c|c|c|c|c|c}
    \toprule
    Dataset & Task & Clean & BackPat & BPP & Wanet & SingTL & MultTL \\
        \midrule
COCO & IR & 75.13 & 74.99 &73.94 &74.54 &74.68 &74.72 \\
 & TR & 80.62 & 81.58 & 77.44 & 78.74 & 80.16 & 81.12 \\
       \midrule
Flickr30k & IR & 59.68 & 59.13 & 59.86 & 61.08 & 60.92 & 57.41 \\
 & TR & 68.37 &68.07 &68.79 &69.86 &71.06&68.14 \\   
         \bottomrule
    \end{tabular}
    }
    \caption{Model utility between \textit{clean} model and other backdoored/poisoned models (CL) (Recall@10). Similar to \ref{tab:defense model utility}.}
    \label{tab:poisoned model utility}
    \vspace{-1em}
\end{table}

\textbf{Utility evaluation.}  We evaluate model utility for image-caption retrieval. \ref{tab:poisoned model utility} shows the performance (Recall@10) of the poisoned model on each attack type as well as the clean model on the test data. We observe that the utility of the poisoned model is at the same level or slightly less than the clean model e.g.~~ BPP in COCO dataset. This implies that despite being trained on poisoned data, models maintain their performance.
We show the model utility after being defended with Semantic Shield and its variants (CL + KE, CL + Attention, weighted CL + Attention) in \ref{tab:defense model utility}.
We largely observe a similar utility compared to the models from \ref{tab:poisoned model utility}. On the Flickr30k dataset, single target or multiple target attack scenario, for TR task, the utility is slightly less than the clean model (\ref{tab:poisoned model utility}, \ref{tab:defense model utility}).

\subsection{Ablations}
\label{sec:ablation}
\textbf{Poisoning rate.} We compare the performance of poisoning attacks at different poisoning rates on three backdoor attacks. We conduct these attacks against the victim model with four different poisoning rates (0.001 to 0.01\%) on the COCO dataset (\ref{fig:ablation_poison_rate}). We observe that attack performance significantly improves with increased poisoning rate, even though the rate is quite low, which demonstrates the vulnerability of contrastively trained VL models to attacks.

\begin{figure}[t]
\centering
\begin{minipage}[t]{0.3\columnwidth}
  \includegraphics[width=\linewidth]{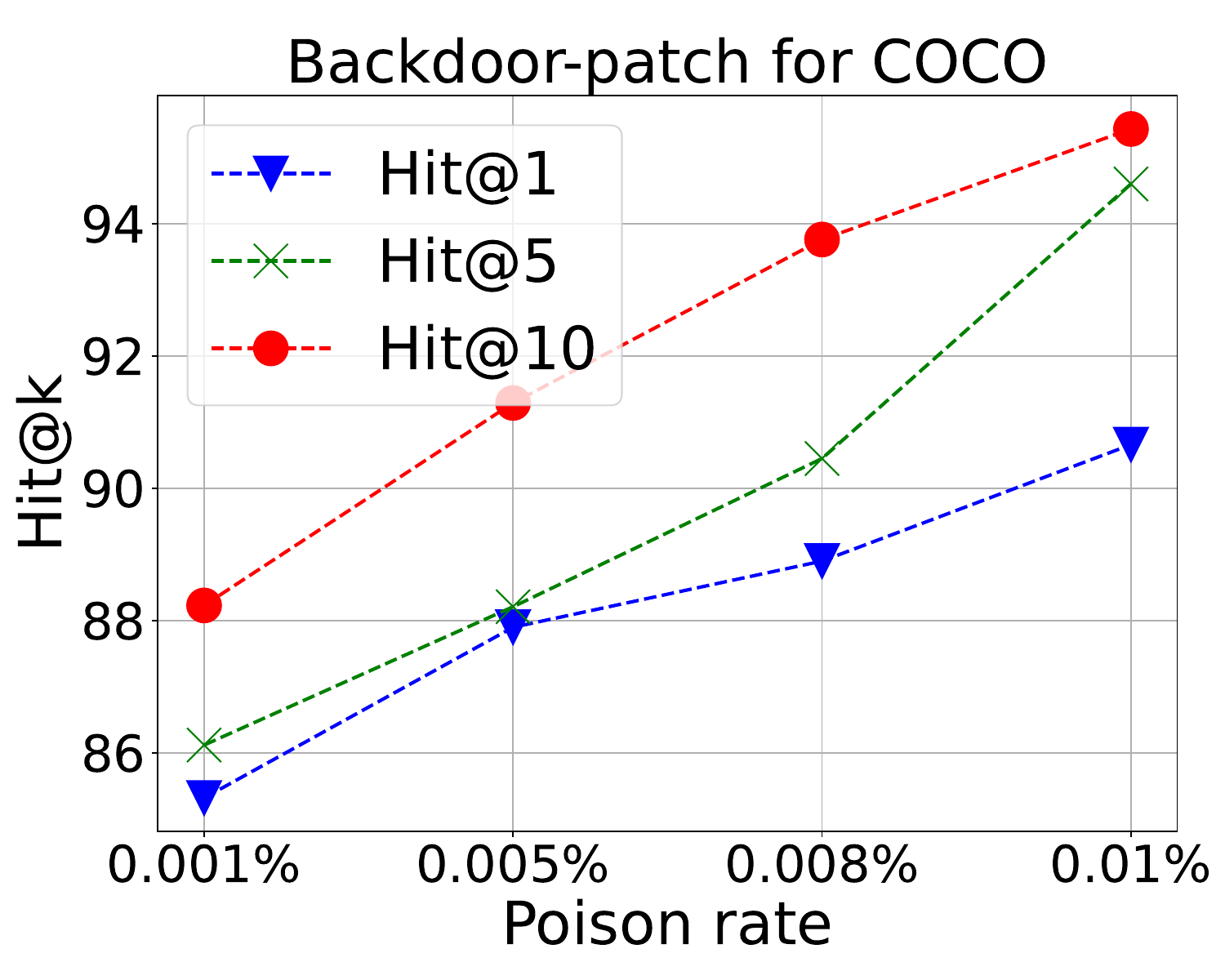}
  \subcaption{Backdoor patch}
   \label{fig:patch_poison_rate}
\end{minipage} 
\begin{minipage}[t]{0.3\columnwidth}
  \includegraphics[width=\linewidth]{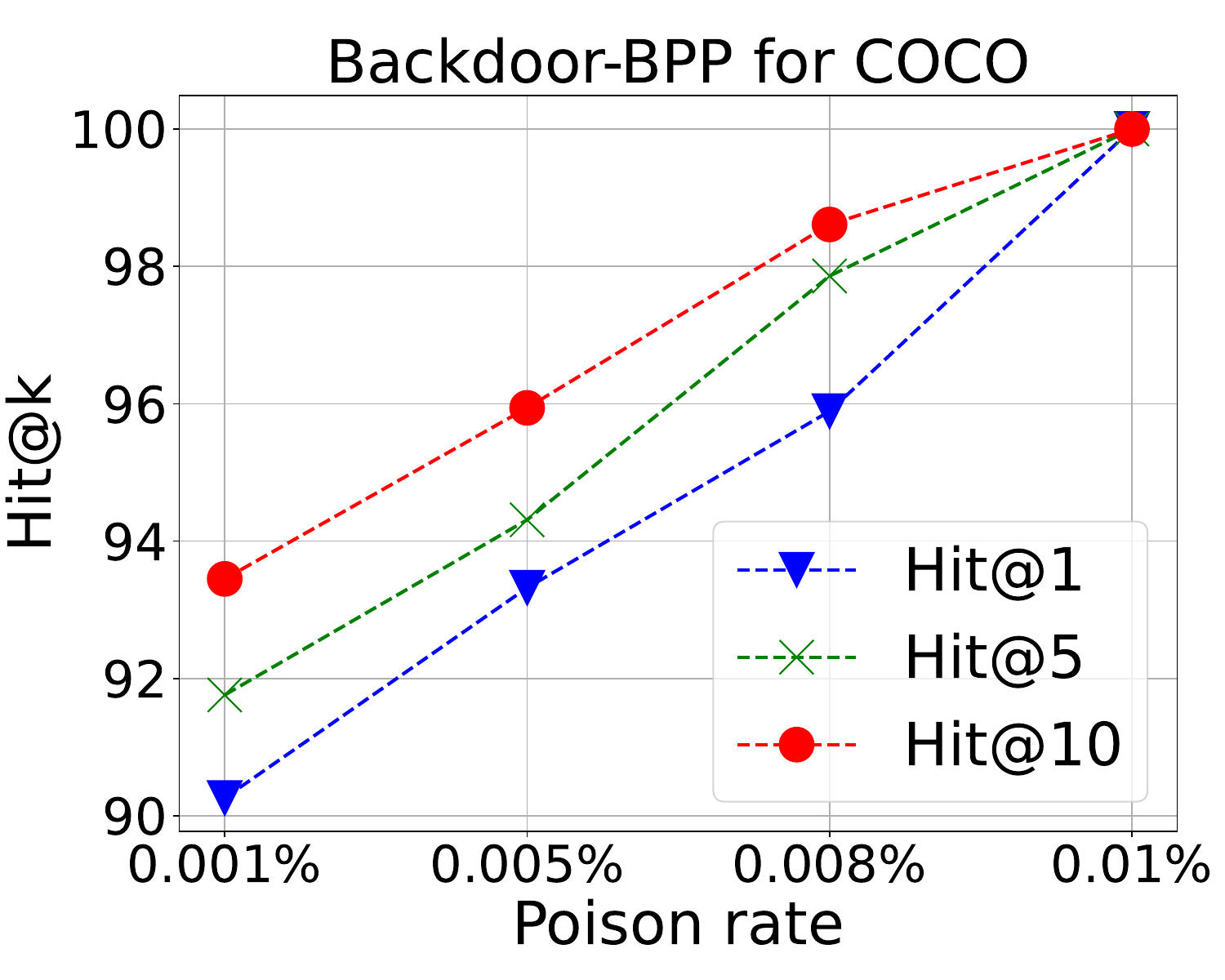}
  \subcaption{BPP}
  \label{fig:bpp_poison_rate}
\end{minipage} 
\begin{minipage}[t]{0.3\columnwidth}
  \includegraphics[width=\linewidth]{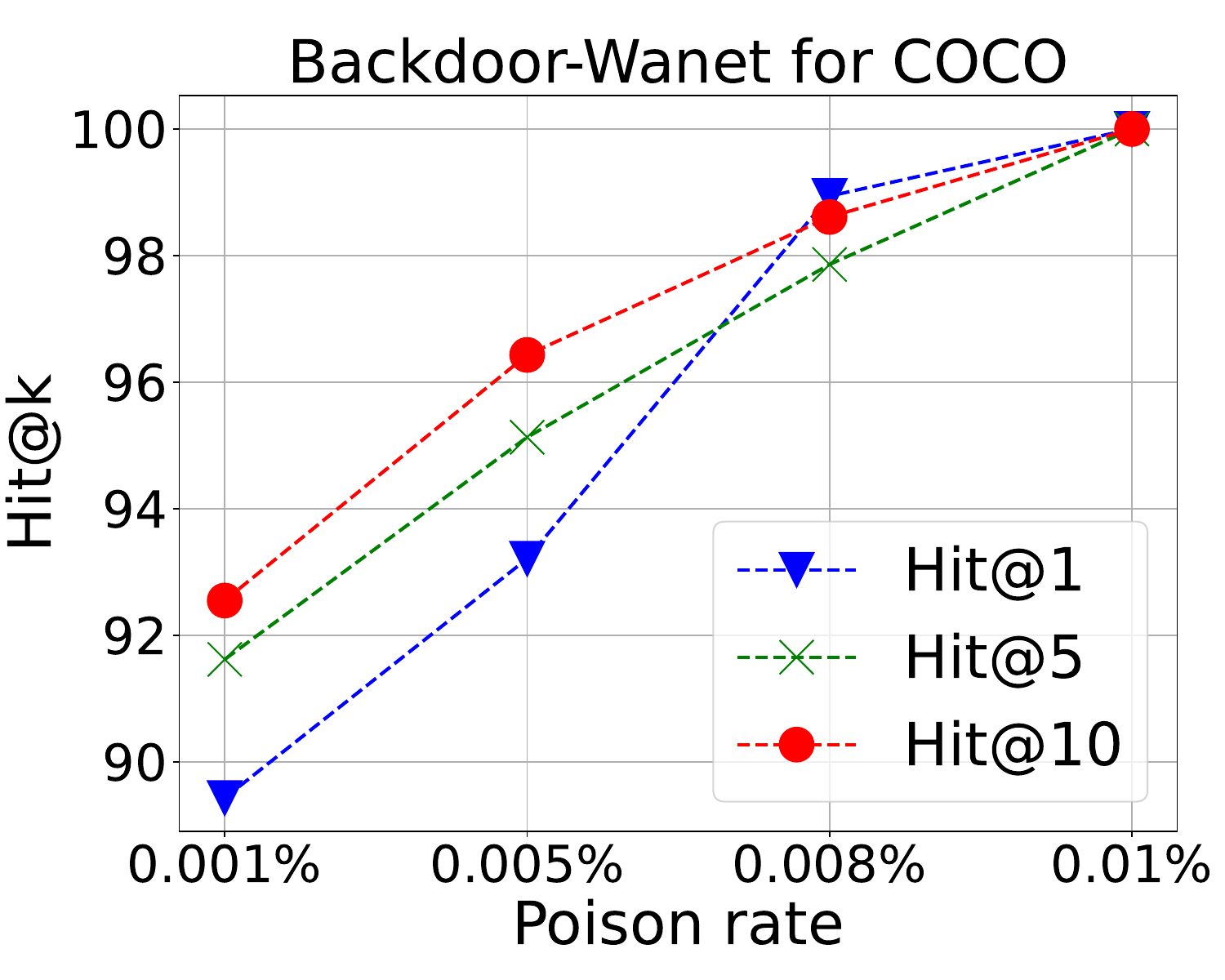}
  \subcaption{Wanet}
    \label{fig:wanet_poison_rate}
\end{minipage}
\caption{Hit@k vs. poisoning rate on backdoored images.}
\vspace{-2em}
\label{fig:ablation_poison_rate}
\end{figure}
\textbf{Fine-tuning epoch.}
In \ref{fig:ablation_finetune_epoch} we use the max poisoning rate ($0.01\%$) from \ref{fig:ablation_poison_rate} to illustrate Semantic Shield's performance at different epochs on the same backdoored samples. We notice that Hit@k gradually reduces for all three attacks, demonstrating the increasing effectiveness of Semantic Shield's defense with increased training.


\begin{figure}[t]
\centering
\vspace{-1em}
\begin{minipage}[t]{0.3\columnwidth}
  \includegraphics[width=\linewidth]{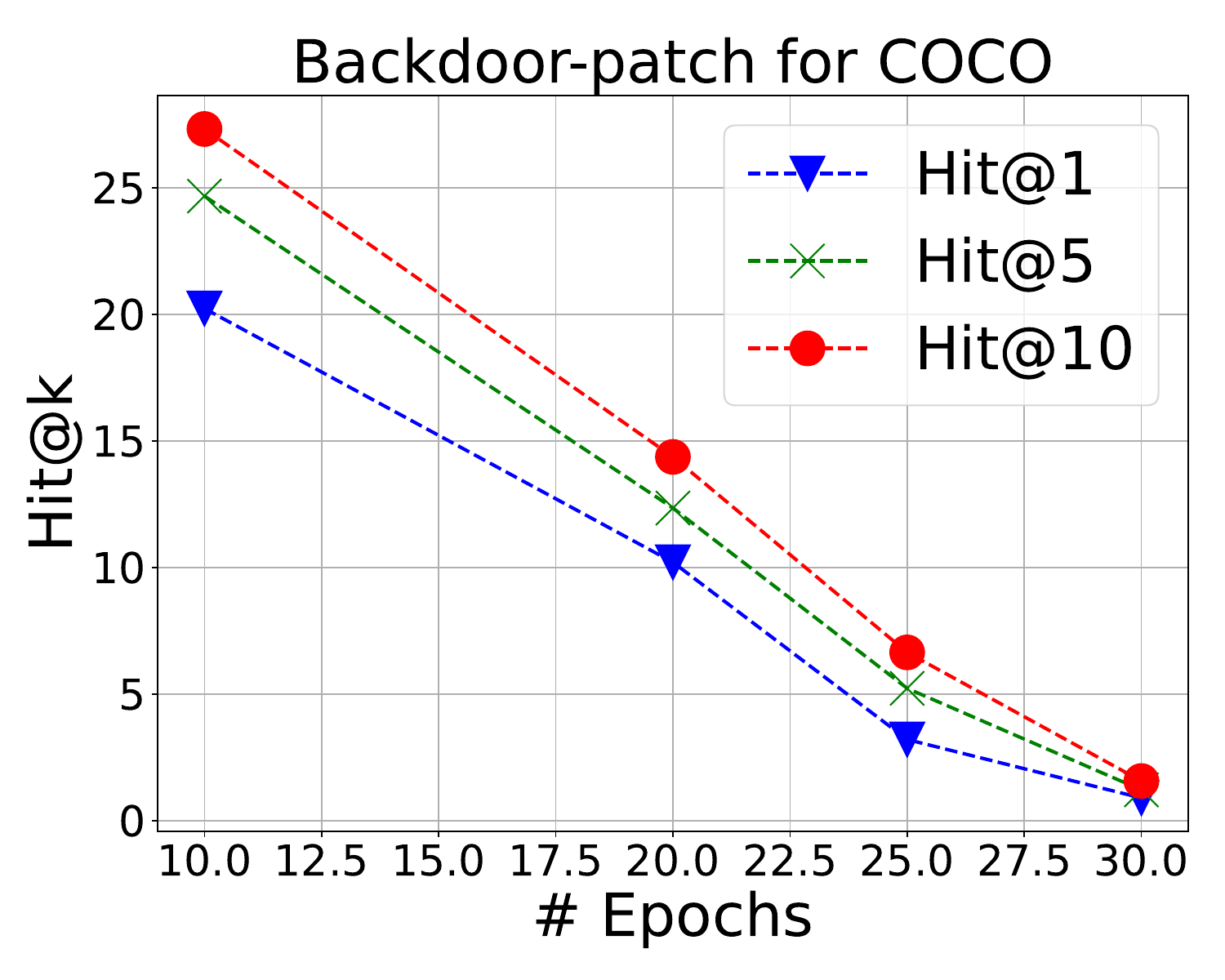}
  \subcaption{Backdoor patch}
   \label{fig:patch_epoch_hit}
\end{minipage} 
\begin{minipage}[t]{0.3\columnwidth}
  \includegraphics[width=\linewidth]{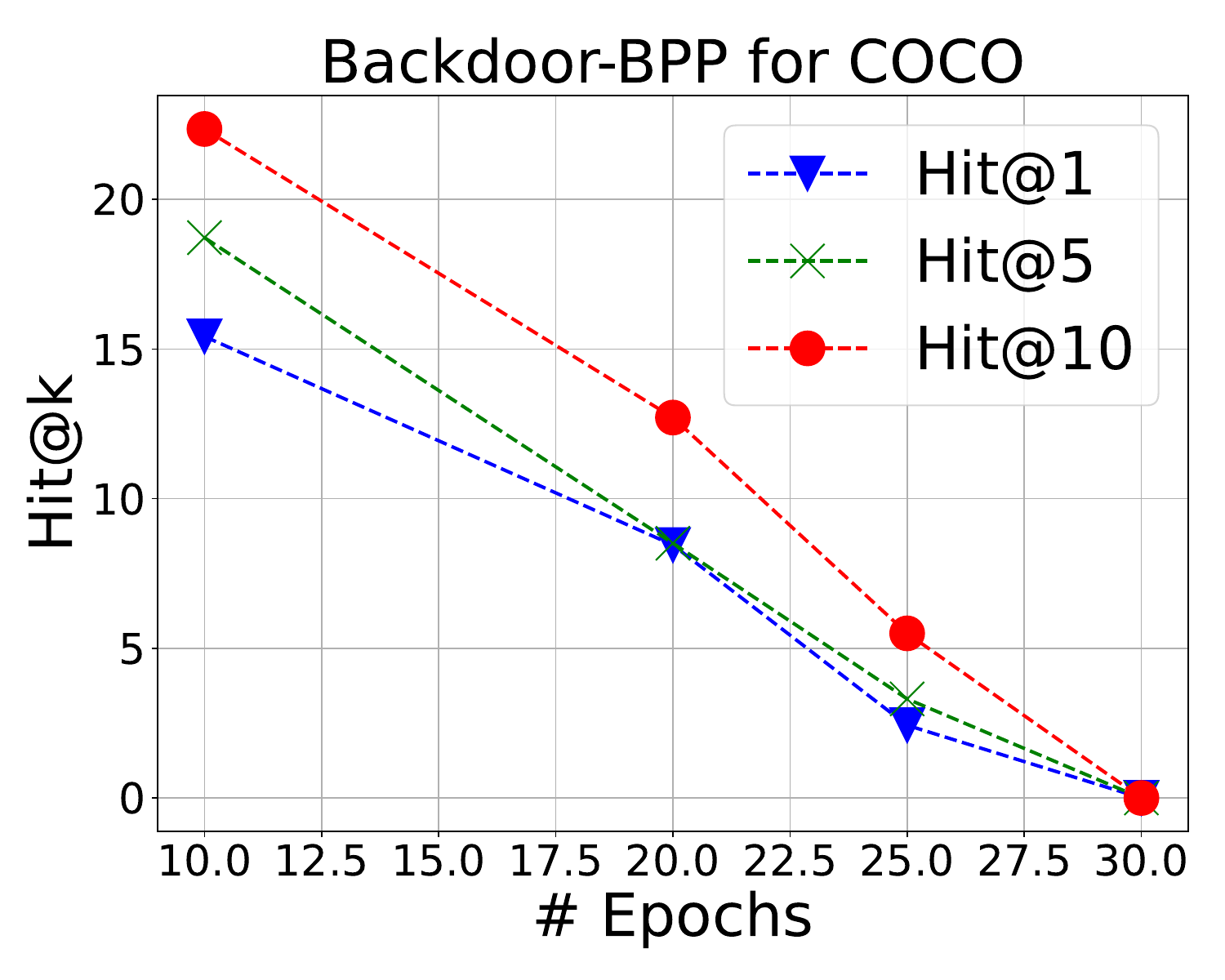}
  \subcaption{BPP}
  \label{fig:bpp_epoch_hit}
\end{minipage} 
\begin{minipage}[t]{0.3\columnwidth}
  \includegraphics[width=\linewidth]{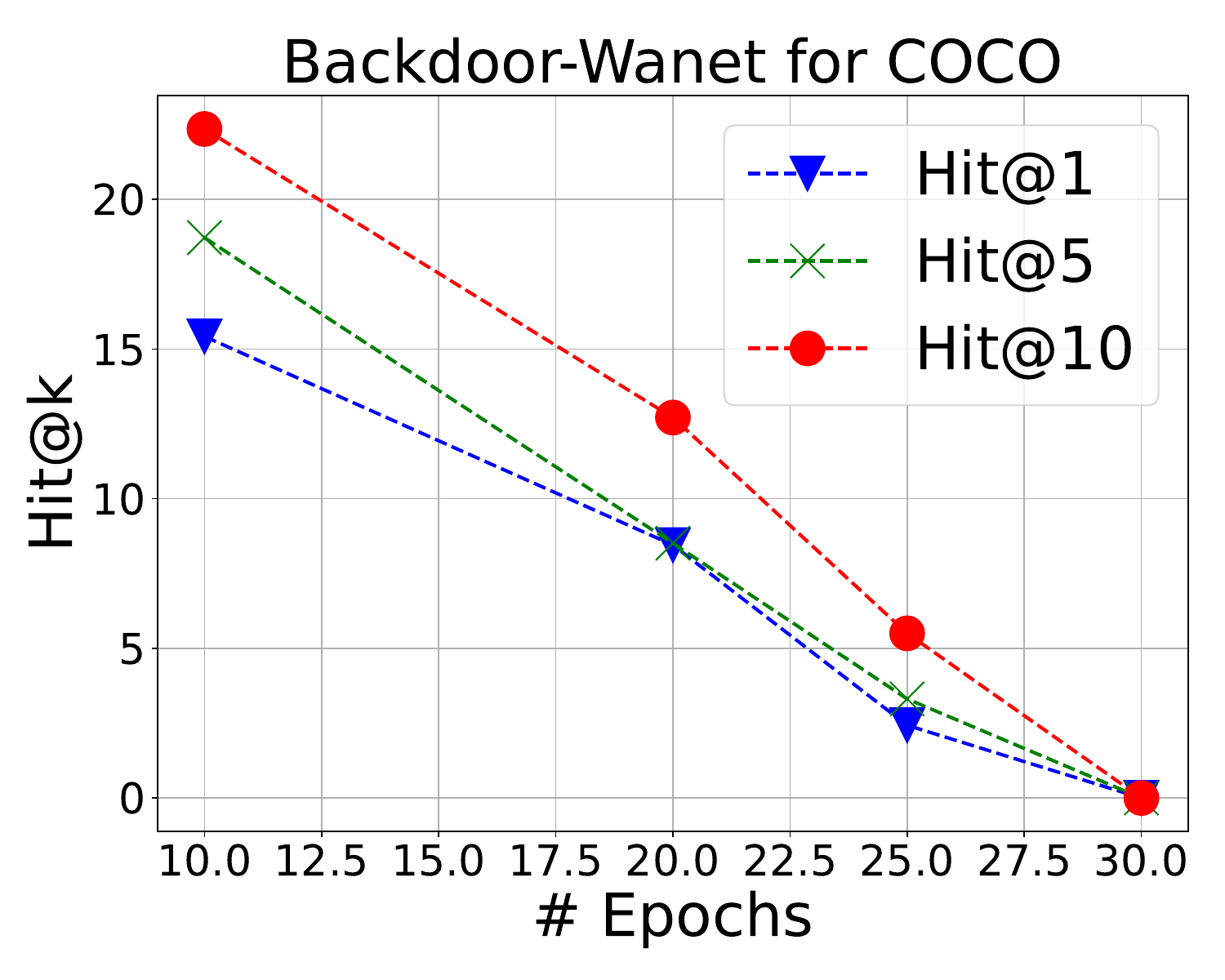}
  \subcaption{Wanet}
    \label{fig:wanet_epoch_hit}
\end{minipage}
\caption{Hit@k vs training epoch for Semantic Shield.}
\vspace{-1em}
\label{fig:ablation_finetune_epoch}
\end{figure}

\section{Qualitative analysis}
\label{qual analysis}
In \ref{fig:qual}, we present the contrast between a model defended by Semantic Shield and an undefended model's attention map. \ref{fig:patch_map} shows that poisoned model pays attention to the patch (bottom right corner). In contrast, the defended model \ref{fig:patch_model}  does not pay any attention to the patch. Next, in \ref{fig:bpp_image} and \ref{fig:wanet_image} two imperceptible noises are injected e.g.~ BPP, Wanet. We wanted to see what happens if we inject the noise randomly throughout the entire images. Poisoned models in \ref{fig:bpp_map} and \ref{fig:wanet_map} show spurious visual signals all over the image. However, our proposed models filters out the noisy signals and defends against poisoning. 

\begin{figure}
\vspace{-1em}
\centering
\begin{minipage}[t]{0.3\columnwidth}
  \includegraphics[width=\linewidth]{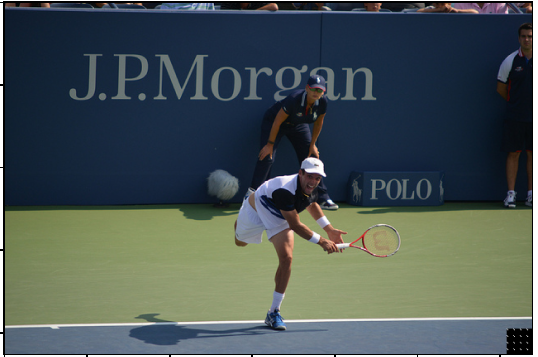}
  \subcaption{Backdoor image with patch bottom right corner}
   \label{fig:patch_image}
\end{minipage} 
\begin{minipage}[t]{0.3\columnwidth}
  \includegraphics[width=\linewidth]{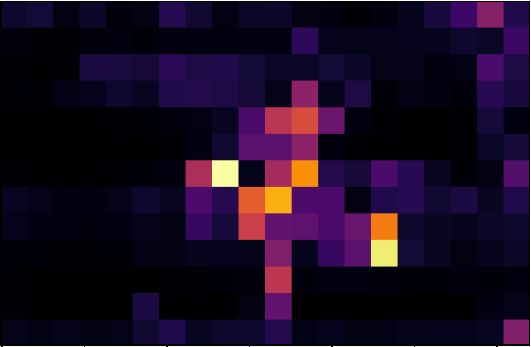}
  \subcaption{Attention map for poisoned model}
  \label{fig:patch_map}
\end{minipage} 
\begin{minipage}[t]{0.3\columnwidth}
  \includegraphics[width=\linewidth]{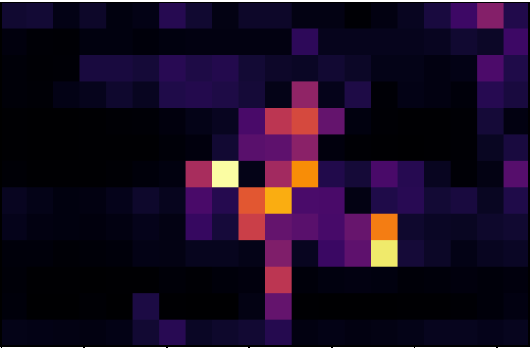}
  \subcaption{Attention map for best model}
    \label{fig:patch_model}
\end{minipage}
\begin{minipage}[t]{0.3\columnwidth}
  \includegraphics[width=\linewidth]{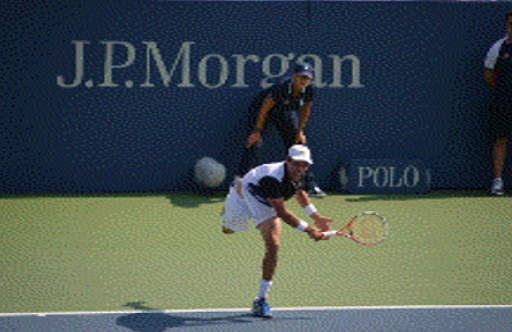}
  \subcaption{Backdoor image with imperceptible noise: BPP}
   \label{fig:bpp_image}
\end{minipage} 
\begin{minipage}[t]{0.3\columnwidth}
  \includegraphics[width=\linewidth]{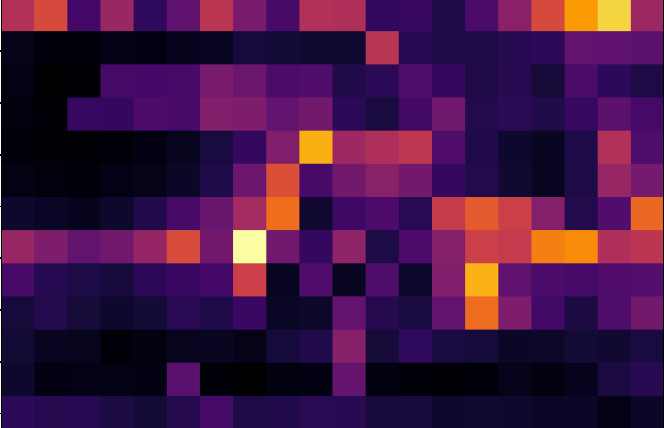}
  \subcaption{Attention map for poisoned model}
   \label{fig:bpp_map}
\end{minipage} 
\begin{minipage}[t]{0.3\columnwidth}
  \includegraphics[width=\linewidth]{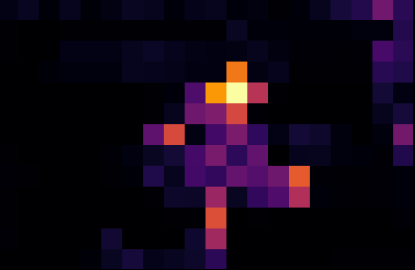}
  \subcaption{Attention map for best model}
   \label{fig:bpp_model}
\end{minipage}
\begin{minipage}[t]{0.3\columnwidth}
  \includegraphics[width=\linewidth]{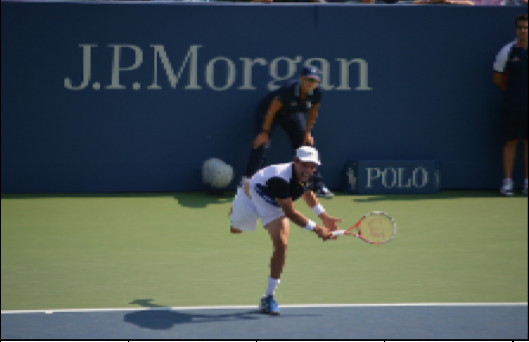}
  \subcaption{Backdoor image with imperceptible noise: Wanet}
  \label{fig:wanet_image}
\end{minipage} 
\begin{minipage}[t]{0.3\columnwidth}
  \includegraphics[width=\linewidth]{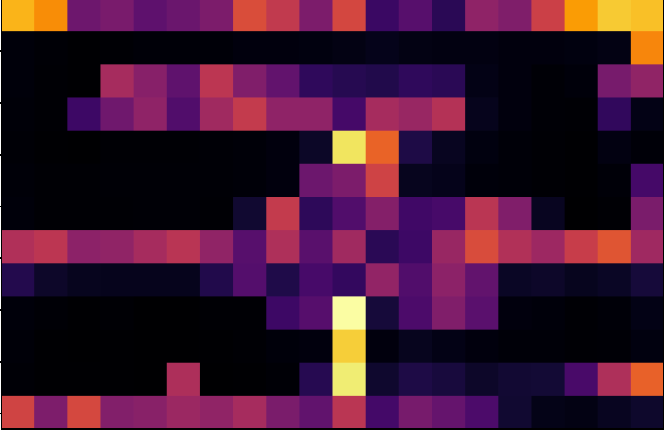}
  \subcaption{Attention map for poisoned model}
  \label{fig:wanet_map}
\end{minipage} 
\begin{minipage}[t]{0.3\columnwidth}
  \includegraphics[width=\linewidth]{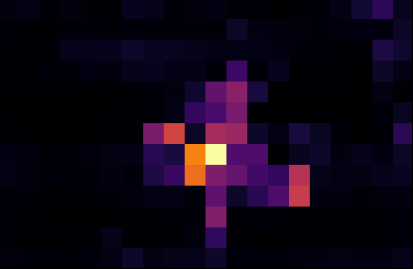}
  \subcaption{Attention map for best model}
\label{fig:wanet_model}
\end{minipage}
\caption{Attention map comparison between our model (weighted CL + attention) and backdoored models for three backdoor attacks.}
\label{fig:qual}

\end{figure}

\section{Conclusion}
\label{sec:conclusion}
In this paper, we introduced Semantic Shield, an approach for defending against attacks on contrastively trained VL models. Our approach works by leveraging external knowledge to guide the model's attention to non-attacked visual regions and samples. We evaluated Semantic Shield against recent backdooring and poisoning attacks and defenses on two benchmarks. Our experiments show that Semantic Shield substantially outperforms existing defenses across all settings. In future work, we will explore a tighter integration of the LLM using prompting by dynamically producing KEs online based on the defended model's current state. In addition, we will explore how multimodal large language models could be used to extract more relevant KEs. While Semantic Shield is successful at defending against attacks on natural images for which there is a meaningful visual-KE alignment, it may be less successful for images such as charts or more abstract text for which clear KEs cannot be extracted. Moreover, it does not preclude the possibility of attacks against the language model via the caption. Future work should explore how the LLM can be jointly defended.
\section{Acknowledgement}
\label{ack}
The authors acknowledge Advanced Research Computing at Virginia Tech for providing computational resources and technical support that have contributed to the results reported within this paper. URL: \url{https://arc.vt.edu/}

{
\small
\bibliographystyle{ieeenat_fullname}

\begin{thebibliography}{67}
\providecommand{\natexlab}[1]{#1}
\providecommand{\url}[1]{\texttt{#1}}
\expandafter\ifx\csname urlstyle\endcsname\relax
  \providecommand{\doi}[1]{doi: #1}\else
  \providecommand{\doi}{doi: \begingroup \urlstyle{rm}\Url}\fi

\bibitem[Akbari et~al.(2021)Akbari, Yuan, Qian, Chuang, Chang, Cui, and Gong]{video_gen}
Hassan Akbari, Liangzhe Yuan, Rui Qian, Wei{-}Hong Chuang, Shih{-}Fu Chang, Yin Cui, and Boqing Gong.
\newblock {VATT:} transformers for multimodal self-supervised learning from raw video, audio and text.
\newblock In \emph{Advances in Neural Information Processing Systems 34: Annual Conference on Neural Information Processing Systems 2021, NeurIPS 2021, December 6-14, 2021, virtual}, pages 24206--24221, 2021.

\bibitem[Bansal et~al.(2023{\natexlab{a}})Bansal, Singhi, Yang, Yin, Grover, and Chang]{bansal2023cleanclip}
Hritik Bansal, Nishad Singhi, Yu Yang, Fan Yin, Aditya Grover, and Kai-Wei Chang.
\newblock Cleanclip: Mitigating data poisoning attacks in multimodal contrastive learning.
\newblock In \emph{ICLR 2023 Workshop on Trustworthy and Reliable Large-Scale Machine Learning Models}, 2023{\natexlab{a}}.

\bibitem[Bansal et~al.(2023{\natexlab{b}})Bansal, Singhi, Yang, Yin, Grover, and Chang]{cleanclip}
Hritik Bansal, Nishad Singhi, Yu Yang, Fan Yin, Aditya Grover, and Kai-Wei Chang.
\newblock Cleanclip: Mitigating data poisoning attacks in multimodal contrastive learning.
\newblock In \emph{Proceedings of the IEEE/CVF International Conference on Computer Vision (ICCV)}, pages 112--123, 2023{\natexlab{b}}.

\bibitem[Biggio et~al.(2012)Biggio, Nelson, and Laskov]{biggio2012poisoning}
Battista Biggio, Blaine Nelson, and Pavel Laskov.
\newblock Poisoning attacks against support vector machines.
\newblock In \emph{Proceedings of the 29th International Coference on International Conference on Machine Learning}, pages 1467--1474, 2012.

\bibitem[Biggio et~al.(2013{\natexlab{a}})Biggio, Pillai, Rota~Bul\`{o}, Ariu, Pelillo, and Roli]{ai_sec}
Battista Biggio, Ignazio Pillai, Samuel Rota~Bul\`{o}, Davide Ariu, Marcello Pelillo, and Fabio Roli.
\newblock Is data clustering in adversarial settings secure?
\newblock In \emph{Proceedings of the 2013 ACM Workshop on Artificial Intelligence and Security}, page 87–98, New York, NY, USA, 2013{\natexlab{a}}. Association for Computing Machinery.

\bibitem[Biggio et~al.(2013{\natexlab{b}})Biggio, Pillai, Rota~Bul{\`o}, Ariu, Pelillo, and Roli]{biggio2013data}
Battista Biggio, Ignazio Pillai, Samuel Rota~Bul{\`o}, Davide Ariu, Marcello Pelillo, and Fabio Roli.
\newblock Is data clustering in adversarial settings secure?
\newblock In \emph{Proceedings of the 2013 ACM workshop on Artificial intelligence and security}, pages 87--98, 2013{\natexlab{b}}.

\bibitem[Carlini and Terzis(2022)]{CarliniT22}
Nicholas Carlini and Andreas Terzis.
\newblock Poisoning and backdooring contrastive learning.
\newblock In \emph{The Tenth International Conference on Learning Representations, {ICLR} 2022, Virtual Event, April 25-29, 2022}. OpenReview.net, 2022.

\bibitem[Chen et~al.(2019)Chen, Carvalho, Baracaldo, Ludwig, Edwards, Lee, Molloy, and Srivastava]{chen2019detecting}
Bryant Chen, Wilka Carvalho, Nathalie Baracaldo, Heiko Ludwig, Benjamin Edwards, Taesung Lee, Ian Molloy, and Biplav Srivastava.
\newblock Detecting backdoor attacks on deep neural networks by activation clustering.
\newblock In \emph{Workshop on Artificial Intelligence Safety}. CEUR-WS, 2019.

\bibitem[Chen et~al.(2021)Chen, Zhang, Zhang, Wang, and Liu]{chen2021pois}
Jian Chen, Xuxin Zhang, Rui Zhang, Chen Wang, and Ling Liu.
\newblock De-pois: An attack-agnostic defense against data poisoning attacks.
\newblock \emph{IEEE Transactions on Information Forensics and Security}, 16:\penalty0 3412--3425, 2021.

\bibitem[Chiang et~al.()Chiang, Li, Lin, Sheng, Wu, Zhang, Zheng, Zhuang, Zhuang, Gonzalez, et~al.]{vicuna}
Wei-Lin Chiang, Zhuohan Li, Zi Lin, Ying Sheng, Zhanghao Wu, Hao Zhang, Lianmin Zheng, Siyuan Zhuang, Yonghao Zhuang, Joseph~E Gonzalez, et~al.
\newblock Vicuna: An open-source chatbot impressing gpt-4 with 90\%* chatgpt quality. 2023.
\newblock \emph{URL https://lmsys. org/blog/2023-03-30-vicuna}, 1\penalty0 (2):\penalty0 3.

\bibitem[Devlin et~al.(2019)Devlin, Chang, Lee, and Toutanova]{bert}
Jacob Devlin, Ming{-}Wei Chang, Kenton Lee, and Kristina Toutanova.
\newblock {BERT:} pre-training of deep bidirectional transformers for language understanding.
\newblock In \emph{Proceedings of the 2019 Conference of the North American Chapter of the Association for Computational Linguistics: Human Language Technologies, {NAACL-HLT} 2019, Minneapolis, MN, USA, June 2-7, 2019, Volume 1 (Long and Short Papers)}, pages 4171--4186. Association for Computational Linguistics, 2019.

\bibitem[Doan et~al.(2021{\natexlab{a}})Doan, Lao, and Li]{doan2021backdoor}
Khoa Doan, Yingjie Lao, and Ping Li.
\newblock Backdoor attack with imperceptible input and latent modification.
\newblock \emph{Advances in Neural Information Processing Systems}, 34:\penalty0 18944--18957, 2021{\natexlab{a}}.

\bibitem[Doan et~al.(2021{\natexlab{b}})Doan, Lao, Zhao, and Li]{doan2021lira}
Khoa Doan, Yingjie Lao, Weijie Zhao, and Ping Li.
\newblock Lira: Learnable, imperceptible and robust backdoor attacks.
\newblock In \emph{Proceedings of the IEEE/CVF international conference on computer vision}, pages 11966--11976, 2021{\natexlab{b}}.

\bibitem[Dorbala et~al.(2022)Dorbala, Sigurdsson, Thomason, Piramuthu, and Sukhatme]{dorbala2022clip}
Vishnu~Sashank Dorbala, Gunnar~A Sigurdsson, Jesse Thomason, Robinson Piramuthu, and Gaurav~S Sukhatme.
\newblock Clip-nav: Using clip for zero-shot vision-and-language navigation.
\newblock In \emph{Workshop on Language and Robotics at CoRL 2022}, 2022.

\bibitem[Feng et~al.(2014)Feng, Wang, and Li]{feng2014cross}
Fangxiang Feng, Xiaojie Wang, and Ruifan Li.
\newblock Cross-modal retrieval with correspondence autoencoder.
\newblock In \emph{Proceedings of the 22nd ACM international conference on Multimedia}, pages 7--16, 2014.

\bibitem[Gonz{\'a}lez-Pizarro and Zannettou(2023)]{gonzalez2023understanding}
Felipe Gonz{\'a}lez-Pizarro and Savvas Zannettou.
\newblock Understanding and detecting hateful content using contrastive learning.
\newblock In \emph{Proceedings of the International AAAI Conference on Web and Social Media}, pages 257--268, 2023.

\bibitem[Gu et~al.(2019)Gu, Liu, Dolan-Gavitt, and Garg]{gu2019badnets}
Tianyu Gu, Kang Liu, Brendan Dolan-Gavitt, and Siddharth Garg.
\newblock Badnets: Evaluating backdooring attacks on deep neural networks.
\newblock \emph{IEEE Access}, 7:\penalty0 47230--47244, 2019.

\bibitem[Hayase et~al.(2021)Hayase, Kong, Somani, and Oh]{hayase2021spectre}
Jonathan Hayase, Weihao Kong, Raghav Somani, and Sewoong Oh.
\newblock Spectre: Defending against backdoor attacks using robust statistics.
\newblock In \emph{International Conference on Machine Learning}, pages 4129--4139. PMLR, 2021.

\bibitem[Huang et~al.(2023)Huang, Mees, Zeng, and Burgard]{huang2023visual}
Chenguang Huang, Oier Mees, Andy Zeng, and Wolfram Burgard.
\newblock Visual language maps for robot navigation.
\newblock In \emph{2023 IEEE International Conference on Robotics and Automation (ICRA)}, pages 10608--10615. IEEE, 2023.

\bibitem[Huang et~al.(2021)Huang, Li, Wu, Qin, and Ren]{huang2021backdoor}
Kunzhe Huang, Yiming Li, Baoyuan Wu, Zhan Qin, and Kui Ren.
\newblock Backdoor defense via decoupling the training process.
\newblock In \emph{International Conference on Learning Representations}, 2021.

\bibitem[Jia et~al.(2021)Jia, Yang, Xia, Chen, Parekh, Pham, Le, Sung, Li, and Duerig]{jia2021scaling}
Chao Jia, Yinfei Yang, Ye Xia, Yi-Ting Chen, Zarana Parekh, Hieu Pham, Quoc Le, Yun-Hsuan Sung, Zhen Li, and Tom Duerig.
\newblock Scaling up visual and vision-language representation learning with noisy text supervision.
\newblock In \emph{International conference on machine learning}, pages 4904--4916. PMLR, 2021.

\bibitem[Kloft and Laskov(2010)]{kloft2010online}
Marius Kloft and Pavel Laskov.
\newblock Online anomaly detection under adversarial impact.
\newblock In \emph{Proceedings of the thirteenth international conference on artificial intelligence and statistics}, pages 405--412. JMLR Workshop and Conference Proceedings, 2010.

\bibitem[Koh and Liang(2017)]{koh2017understanding}
Pang~Wei Koh and Percy Liang.
\newblock Understanding black-box predictions via influence functions.
\newblock In \emph{International conference on machine learning}, pages 1885--1894. PMLR, 2017.

\bibitem[Li et~al.(2023)Li, Pang, Xi, Du, Ji, Yao, and Wang]{li2023embarrassingly}
Changjiang Li, Ren Pang, Zhaohan Xi, Tianyu Du, Shouling Ji, Yuan Yao, and Ting Wang.
\newblock An embarrassingly simple backdoor attack on self-supervised learning.
\newblock In \emph{Proceedings of the IEEE/CVF International Conference on Computer Vision}, pages 4367--4378, 2023.

\bibitem[Li et~al.(2021{\natexlab{a}})Li, Selvaraju, Gotmare, Joty, Xiong, and Hoi]{li2021align}
Junnan Li, Ramprasaath Selvaraju, Akhilesh Gotmare, Shafiq Joty, Caiming Xiong, and Steven Chu~Hong Hoi.
\newblock Align before fuse: Vision and language representation learning with momentum distillation.
\newblock \emph{Advances in neural information processing systems}, 34:\penalty0 9694--9705, 2021{\natexlab{a}}.

\bibitem[Li et~al.(2022)Li, Li, Xiong, and Hoi]{img-gen_2}
Junnan Li, Dongxu Li, Caiming Xiong, and Steven C.~H. Hoi.
\newblock {BLIP:} bootstrapping language-image pre-training for unified vision-language understanding and generation.
\newblock In \emph{International Conference on Machine Learning, {ICML} 2022, 17-23 July 2022, Baltimore, Maryland, {USA}}, pages 12888--12900. {PMLR}, 2022.

\bibitem[Li et~al.(2021{\natexlab{b}})Li, Lyu, Koren, Lyu, Li, and Ma]{abl}
Yige Li, Xixiang Lyu, Nodens Koren, Lingjuan Lyu, Bo Li, and Xingjun Ma.
\newblock Anti-backdoor learning: Training clean models on poisoned data.
\newblock In \emph{Advances in Neural Information Processing Systems 34: Annual Conference on Neural Information Processing Systems 2021, NeurIPS 2021, December 6-14, 2021, virtual}, pages 14900--14912, 2021{\natexlab{b}}.

\bibitem[Lin et~al.(2014)Lin, Maire, Belongie, Hays, Perona, Ramanan, Doll{\'a}r, and Zitnick]{coco}
Tsung-Yi Lin, Michael Maire, Serge Belongie, James Hays, Pietro Perona, Deva Ramanan, Piotr Doll{\'a}r, and C~Lawrence Zitnick.
\newblock Microsoft coco: Common objects in context.
\newblock In \emph{Computer Vision--ECCV 2014: 13th European Conference, Zurich, Switzerland, September 6-12, 2014, Proceedings, Part V 13}, pages 740--755. Springer, 2014.

\bibitem[Liu et~al.(2023)Liu, Sangiovanni-Vincentelli, and Yue]{liu2023beating}
Min Liu, Alberto Sangiovanni-Vincentelli, and Xiangyu Yue.
\newblock Beating backdoor attack at its own game.
\newblock In \emph{Proceedings of the IEEE/CVF International Conference on Computer Vision}, pages 4620--4629, 2023.

\bibitem[Liu et~al.(2022)Liu, Fan, Chen, Liu, Ma, Wang, and Ma]{liu2022backdoor}
Yang Liu, Mingyuan Fan, Cen Chen, Ximeng Liu, Zhuo Ma, Li Wang, and Jianfeng Ma.
\newblock Backdoor defense with machine unlearning.
\newblock In \emph{IEEE INFOCOM 2022-IEEE Conference on Computer Communications}, pages 280--289. IEEE, 2022.

\bibitem[Majumdar et~al.(2022)Majumdar, Aggarwal, Devnani, Hoffman, and Batra]{majumdar2022zson}
Arjun Majumdar, Gunjan Aggarwal, Bhavika Devnani, Judy Hoffman, and Dhruv Batra.
\newblock Zson: Zero-shot object-goal navigation using multimodal goal embeddings.
\newblock \emph{Advances in Neural Information Processing Systems}, 35:\penalty0 32340--32352, 2022.

\bibitem[Menon and Vondrick(2023)]{carl2023}
Sachit Menon and Carl Vondrick.
\newblock Visual classification via description from large language models.
\newblock In \emph{The Eleventh International Conference on Learning Representations, {ICLR} 2023, Kigali, Rwanda, May 1-5, 2023}. OpenReview.net, 2023.

\bibitem[Nguyen and Tran(2021)]{wanet}
Tuan~Anh Nguyen and Anh~Tuan Tran.
\newblock Wanet - imperceptible warping-based backdoor attack.
\newblock In \emph{International Conference on Learning Representations}, 2021.

\bibitem[Phan et~al.(2022)Phan, Shi, Xie, Zhang, Li, Zhao, Liu, Wang, Chen, and Yuan]{phan2022ribac}
Huy Phan, Cong Shi, Yi Xie, Tianfang Zhang, Zhuohang Li, Tianming Zhao, Jian Liu, Yan Wang, Yingying Chen, and Bo Yuan.
\newblock Ribac: Towards r obust and i mperceptible b ackdoor a ttack against c ompact dnn.
\newblock In \emph{European Conference on Computer Vision}, pages 708--724. Springer, 2022.

\bibitem[Qiu et~al.(2021)Qiu, Zeng, Guo, Zhang, Qiu, and Thuraisingham]{qiu2021deepsweep}
Han Qiu, Yi Zeng, Shangwei Guo, Tianwei Zhang, Meikang Qiu, and Bhavani Thuraisingham.
\newblock Deepsweep: An evaluation framework for mitigating dnn backdoor attacks using data augmentation.
\newblock In \emph{Proceedings of the 2021 ACM Asia Conference on Computer and Communications Security}, pages 363--377, 2021.

\bibitem[Radford et~al.(2021)Radford, Kim, Hallacy, Ramesh, Goh, Agarwal, Sastry, Askell, Mishkin, Clark, Krueger, and Sutskever]{Radford2021LearningTV}
Alec Radford, Jong~Wook Kim, Chris Hallacy, Aditya Ramesh, Gabriel Goh, Sandhini Agarwal, Girish Sastry, Amanda Askell, Pamela Mishkin, Jack Clark, Gretchen Krueger, and Ilya Sutskever.
\newblock Learning transferable visual models from natural language supervision.
\newblock In \emph{International Conference on Machine Learning}, 2021.

\bibitem[Ramesh et~al.(2022)Ramesh, Dhariwal, Nichol, Chu, and Chen]{img-gen}
Aditya Ramesh, Prafulla Dhariwal, Alex Nichol, Casey Chu, and Mark Chen.
\newblock Hierarchical text-conditional image generation with {CLIP} latents.
\newblock \emph{CoRR}, abs/2204.06125, 2022.

\bibitem[Saha et~al.(2020)Saha, Subramanya, and Pirsiavash]{saha2020hidden}
Aniruddha Saha, Akshayvarun Subramanya, and Hamed Pirsiavash.
\newblock Hidden trigger backdoor attacks.
\newblock In \emph{Proceedings of the AAAI conference on artificial intelligence}, pages 11957--11965, 2020.

\bibitem[Saha et~al.(2022)Saha, Tejankar, Koohpayegani, and Pirsiavash]{saha2022}
Aniruddha Saha, Ajinkya Tejankar, Soroush~Abbasi Koohpayegani, and Hamed Pirsiavash.
\newblock Backdoor attacks on self-supervised learning.
\newblock In \emph{{IEEE/CVF} Conference on Computer Vision and Pattern Recognition, {CVPR} 2022, New Orleans, LA, USA, June 18-24, 2022}, pages 13327--13336. {IEEE}, 2022.

\bibitem[Shafahi et~al.(2018)Shafahi, Huang, Najibi, Suciu, Studer, Dumitras, and Goldstein]{Shafahi2018}
Ali Shafahi, W.~Ronny Huang, Mahyar Najibi, Octavian Suciu, Christoph Studer, Tudor Dumitras, and Tom Goldstein.
\newblock Poison frogs! targeted clean-label poisoning attacks on neural networks.
\newblock In \emph{Proceedings of the 32nd International Conference on Neural Information Processing Systems}, page 6106–6116, Red Hook, NY, USA, 2018. Curran Associates Inc.

\bibitem[Shin et~al.(2022)Shin, Park, Woo, Cho, Oh, and Song]{shin2022clip}
Wonyoung Shin, Jonghun Park, Taekang Woo, Yongwoo Cho, Kwangjin Oh, and Hwanjun Song.
\newblock e-clip: Large-scale vision-language representation learning in e-commerce.
\newblock In \emph{Proceedings of the 31st ACM International Conference on Information \& Knowledge Management}, pages 3484--3494, 2022.

\bibitem[Steiner et~al.(2022)Steiner, Kolesnikov, Zhai, Wightman, Uszkoreit, and Beyer]{imagenet21k}
Andreas Steiner, Alexander Kolesnikov, Xiaohua Zhai, Ross Wightman, Jakob Uszkoreit, and Lucas Beyer.
\newblock How to train your vit? data, augmentation, and regularization in vision transformers.
\newblock \emph{Trans. Mach. Learn. Res.}, 2022, 2022.

\bibitem[Tang et~al.(2021)Tang, Wang, Tang, and Zhang]{tang2021demon}
Di Tang, XiaoFeng Wang, Haixu Tang, and Kehuan Zhang.
\newblock Demon in the variant: Statistical analysis of $\{$DNNs$\}$ for robust backdoor contamination detection.
\newblock In \emph{30th USENIX Security Symposium (USENIX Security 21)}, pages 1541--1558, 2021.

\bibitem[Thomas and Kovashka(2020)]{thomas2020preserving}
Christopher Thomas and Adriana Kovashka.
\newblock Preserving semantic neighborhoods for robust cross-modal retrieval.
\newblock In \emph{Computer Vision--ECCV 2020: 16th European Conference, Glasgow, UK, August 23--28, 2020, Proceedings, Part XVIII 16}, pages 317--335. Springer, 2020.

\bibitem[Tolpegin et~al.(2020)Tolpegin, Truex, Gursoy, and Liu]{tolpegin2020data}
Vale Tolpegin, Stacey Truex, Mehmet~Emre Gursoy, and Ling Liu.
\newblock Data poisoning attacks against federated learning systems.
\newblock In \emph{Computer Security--ESORICS 2020: 25th European Symposium on Research in Computer Security, ESORICS 2020, Guildford, UK, September 14--18, 2020, Proceedings, Part I 25}, pages 480--501. Springer, 2020.

\bibitem[Tran et~al.(2018)Tran, Li, and Madry]{tran2018spectral}
Brandon Tran, Jerry Li, and Aleksander Madry.
\newblock Spectral signatures in backdoor attacks.
\newblock \emph{Advances in neural information processing systems}, 31, 2018.

\bibitem[Tsai et~al.(2022)Tsai, Lin, and Hsieh]{tsai2022generating}
Wei~Lun Tsai, Jacob~J Lin, and Shang-Hsien Hsieh.
\newblock Generating construction safety observations via clip-based image-language embedding.
\newblock In \emph{European Conference on Computer Vision}, pages 366--381. Springer, 2022.

\bibitem[Wang et~al.(2022{\natexlab{a}})Wang, Hong, Zhang, Zhou, and Wang]{wang2022trap}
Haotao Wang, Junyuan Hong, Aston Zhang, Jiayu Zhou, and Zhangyang Wang.
\newblock Trap and replace: Defending backdoor attacks by trapping them into an easy-to-replace subnetwork.
\newblock \emph{Advances in neural information processing systems}, 35:\penalty0 36026--36039, 2022{\natexlab{a}}.

\bibitem[Wang and Chen(2022)]{wang2022improving}
Lin Wang and Jie Chen.
\newblock Improving radiology report generation with adaptive attention.
\newblock In \emph{Multimodal AI in healthcare: A paradigm shift in health intelligence}, pages 293--305. Springer, 2022.

\bibitem[Wang et~al.(2023)Wang, Zhang, Xu, Xu, Xu, and Wang]{wang2023cross}
Longzheng Wang, Chuang Zhang, Hongbo Xu, Yongxiu Xu, Xiaohan Xu, and Siqi Wang.
\newblock Cross-modal contrastive learning for multimodal fake news detection.
\newblock In \emph{Proceedings of the 31st ACM International Conference on Multimedia}, pages 5696--5704, 2023.

\bibitem[Wang et~al.(2022{\natexlab{b}})Wang, Zhai, and Ma]{bpp}
Zhenting Wang, Juan Zhai, and Shiqing Ma.
\newblock Bppattack: Stealthy and efficient trojan attacks against deep neural networks via image quantization and contrastive adversarial learning.
\newblock In \emph{{IEEE/CVF} Conference on Computer Vision and Pattern Recognition, {CVPR} 2022, New Orleans, LA, USA, June 18-24, 2022}, pages 15054--15063. {IEEE}, 2022{\natexlab{b}}.

\bibitem[Weerasinghe et~al.(2021)Weerasinghe, Alpcan, Erfani, and Leckie]{weerasinghe2021defending}
Sandamal Weerasinghe, Tansu Alpcan, Sarah~M Erfani, and Christopher Leckie.
\newblock Defending support vector machines against data poisoning attacks.
\newblock \emph{IEEE Transactions on Information Forensics and Security}, 16:\penalty0 2566--2578, 2021.

\bibitem[Wu and Wang(2021)]{wu2021adversarial}
Dongxian Wu and Yisen Wang.
\newblock Adversarial neuron pruning purifies backdoored deep models.
\newblock \emph{Advances in Neural Information Processing Systems}, 34:\penalty0 16913--16925, 2021.

\bibitem[Xiao et~al.(2015)Xiao, Biggio, Brown, Fumera, Eckert, and Roli]{xiao2015feature}
Huang Xiao, Battista Biggio, Gavin Brown, Giorgio Fumera, Claudia Eckert, and Fabio Roli.
\newblock Is feature selection secure against training data poisoning?
\newblock In \emph{International conference on machine learning}, pages 1689--1698. PMLR, 2015.

\bibitem[Yang et~al.(2023{\natexlab{a}})Yang, Gao, and Mirzasoleiman]{yang2023robust}
Wenhan Yang, Jingdong Gao, and Baharan Mirzasoleiman.
\newblock Robust contrastive language-image pretraining against data poisoning and backdoor attacks.
\newblock In \emph{Thirty-seventh Conference on Neural Information Processing Systems}, 2023{\natexlab{a}}.

\bibitem[Yang et~al.(2023{\natexlab{b}})Yang, He, Li, Backes, Humbert, Berrang, and Zhang]{icml}
Ziqing Yang, Xinlei He, Zheng Li, Michael Backes, Mathias Humbert, Pascal Berrang, and Yang Zhang.
\newblock Data poisoning attacks against multimodal encoders.
\newblock In \emph{International Conference on Machine Learning, {ICML} 2023, 23-29 July 2023, Honolulu, Hawaii, {USA}}, pages 39299--39313. {PMLR}, 2023{\natexlab{b}}.

\bibitem[Yang et~al.(2023{\natexlab{c}})Yang, He, Li, Backes, Humbert, Berrang, and Zhang]{icml_poison}
Ziqing Yang, Xinlei He, Zheng Li, Michael Backes, Mathias Humbert, Pascal Berrang, and Yang Zhang.
\newblock Data poisoning attacks against multimodal encoders.
\newblock In \emph{International Conference on Machine Learning, {ICML} 2023, 23-29 July 2023, Honolulu, Hawaii, {USA}}, pages 39299--39313. {PMLR}, 2023{\natexlab{c}}.

\bibitem[Young et~al.(2014)Young, Lai, Hodosh, and Hockenmaier]{flickr}
Peter Young, Alice Lai, Micah Hodosh, and Julia Hockenmaier.
\newblock From image descriptions to visual denotations: New similarity metrics for semantic inference over event descriptions.
\newblock \emph{Trans. Assoc. Comput. Linguistics}, 2:\penalty0 67--78, 2014.

\bibitem[Yu et~al.(2022)Yu, Wang, Vasudevan, Yeung, Seyedhosseini, and Wu]{yu2022coca}
Jiahui Yu, Zirui Wang, Vijay Vasudevan, Legg Yeung, Mojtaba Seyedhosseini, and Yonghui Wu.
\newblock Coca: Contrastive captioners are image-text foundation models.
\newblock \emph{Transactions on Machine Learning Research}, 2022.

\bibitem[Zeng et~al.(2021)Zeng, Chen, Park, Mao, Jin, and Jia]{zeng2021adversarial}
Yi Zeng, Si Chen, Won Park, Zhuoqing Mao, Ming Jin, and Ruoxi Jia.
\newblock Adversarial unlearning of backdoors via implicit hypergradient.
\newblock In \emph{International Conference on Learning Representations}, 2021.

\bibitem[Zhai et~al.(2022)Zhai, Wang, Mustafa, Steiner, Keysers, Kolesnikov, and Beyer]{zhai2022lit}
Xiaohua Zhai, Xiao Wang, Basil Mustafa, Andreas Steiner, Daniel Keysers, Alexander Kolesnikov, and Lucas Beyer.
\newblock Lit: Zero-shot transfer with locked-image text tuning.
\newblock In \emph{Proceedings of the IEEE/CVF Conference on Computer Vision and Pattern Recognition}, pages 18123--18133, 2022.

\bibitem[Zhang et~al.(2014)Zhang, Yang, Luan, Yang, and Chua]{zhang2014start}
Hanwang Zhang, Yang Yang, Huanbo Luan, Shuicheng Yang, and Tat-Seng Chua.
\newblock Start from scratch: Towards automatically identifying, modeling, and naming visual attributes.
\newblock In \emph{Proceedings of the 22nd ACM international conference on Multimedia}, pages 187--196, 2014.

\bibitem[Zhang et~al.(2022{\natexlab{a}})Zhang, Liu, Jia, and Gong]{zhang2022corruptencoder}
Jinghuai Zhang, Hongbin Liu, Jinyuan Jia, and Neil~Zhenqiang Gong.
\newblock Corruptencoder: Data poisoning based backdoor attacks to contrastive learning.
\newblock \emph{arXiv preprint arXiv:2211.08229}, 2022{\natexlab{a}}.

\bibitem[Zhang and Lu(2018)]{zhang2018deep}
Ying Zhang and Huchuan Lu.
\newblock Deep cross-modal projection learning for image-text matching.
\newblock In \emph{Proceedings of the European conference on computer vision (ECCV)}, pages 686--701, 2018.

\bibitem[Zhang et~al.(2022{\natexlab{b}})Zhang, Jiang, Miura, Manning, and Langlotz]{zhang2022contrastive}
Yuhao Zhang, Hang Jiang, Yasuhide Miura, Christopher~D Manning, and Curtis~P Langlotz.
\newblock Contrastive learning of medical visual representations from paired images and text.
\newblock In \emph{Machine Learning for Healthcare Conference}, pages 2--25. PMLR, 2022{\natexlab{b}}.

\bibitem[Zhao and Lao(2022)]{zhao2022towards}
Bingyin Zhao and Yingjie Lao.
\newblock Towards class-oriented poisoning attacks against neural networks.
\newblock In \emph{Proceedings of the IEEE/CVF Winter Conference on Applications of Computer Vision}, pages 3741--3750, 2022.

\bibitem[Zhou et~al.(2023)Zhou, Yang, Ying, Qian, and Zhang]{zhou2023multimodal}
Yangming Zhou, Yuzhou Yang, Qichao Ying, Zhenxing Qian, and Xinpeng Zhang.
\newblock Multimodal fake news detection via clip-guided learning.
\newblock In \emph{2023 IEEE International Conference on Multimedia and Expo (ICME)}, pages 2825--2830. IEEE, 2023.

\end{thebibliography}

}


\end{document}